%% file: neurips_2026.tex
\documentclass{article}

\usepackage[numbers,sort&compress]{natbib}

 \usepackage[main, final]{neurips_2026}

\usepackage[utf8]{inputenc} 
\usepackage[T1]{fontenc}    
\usepackage{hyperref}       
\usepackage{url}            
\usepackage{booktabs}       
\usepackage{amsfonts}       
\usepackage{nicefrac}       
\usepackage{microtype}      
\usepackage{xcolor}         

\usepackage{amsmath}
\usepackage{graphicx}
\usepackage{tabularx}
\usepackage{wrapfig}
\usepackage{multirow}
\usepackage{etoc}

\title{Invariant Features in Language Models: Geometric Characterization and Model Attribution}

\author{
Agnibh Dasgupta\thanks{Equal contribution.} \quad
Abdullah All Tanvir\footnotemark[1] \quad
Xin Zhong\thanks{Corresponding author: xzhong@unomaha.edu} \\ [0.2em]
Department of Computer Science \\
University of Nebraska Omaha, Omaha, NE, USA \\[0.2em]
}

\begin{document}

\maketitle

\begin{abstract}
Language models exhibit strong robustness to paraphrasing, suggesting that semantic information may be encoded through stable internal representations, yet the structure and origin of such invariance remain unclear. We propose a local geometric framework in which semantically equivalent inputs produce representations that occupy structured regions in latent space, with paraphrastic variation along nuisance directions and semantic identity preserved in invariant subspaces. Building on this view, we make three contributions: (1) a geometric characterization of invariant latent features, (2) a contrastive subspace discovery method that separates semantic-changing from semantic-preserving variation, and (3) an application of invariant representations to zero-shot model attribution. Across models and layers, empirical results support these contributions. Invariant structure emerges in specific depth regions, semantic displacement lies largely outside the nuisance subspace, and representation-level interventions indicate a causal role of invariant components in model outputs. Invariant representations also capture model-specific geometric patterns, enabling accurate attribution. Together, these findings support a local geometric characterization of semantic invariance in language model representations.
\end{abstract}


\etocdepthtag.toc{main}

\section{Introduction}
\label{sec:intro}

\input{sections/introduction}

\section{Related Work}
\label{sec:related}
\input{sections/related}

\section{Proposed Framework for Invariant Latent Feature Analysis}
\label{sec:method}
\input{sections/method}

\section{Experiments}
\label{sec:experiment}
\input{sections/experiment}

\section{Limitation and Discussion}
\label{sec:limitation}

\input{sections/limitation}


\section{Conclusion}
\label{sec:conclusion}
\input{sections/conclusion}

\newpage
\bibliographystyle{IEEEtran}
\bibliography{ref}


\newpage
\input{sections/appendix}

\clearpage



\end{document}

%% file: sections/introduction.tex

Large language models (LLMs) exhibit a striking robustness to surface-level variation. Prompts that differ substantially in wording, style, or syntactic structure often lead to similar internal reasoning and consistent outputs. This behavior suggests that LLMs encode semantic information in a stable internal representation that abstracts away from lexical form. However, despite extensive empirical success, it remains unclear where such stable semantic representations emerge within model architectures, how they can be formally characterized, and how their geometric structure differs across models.

\begin{figure}[h]
    \centering
    \vspace{-0.75em}
    \includegraphics[width=0.9\linewidth]{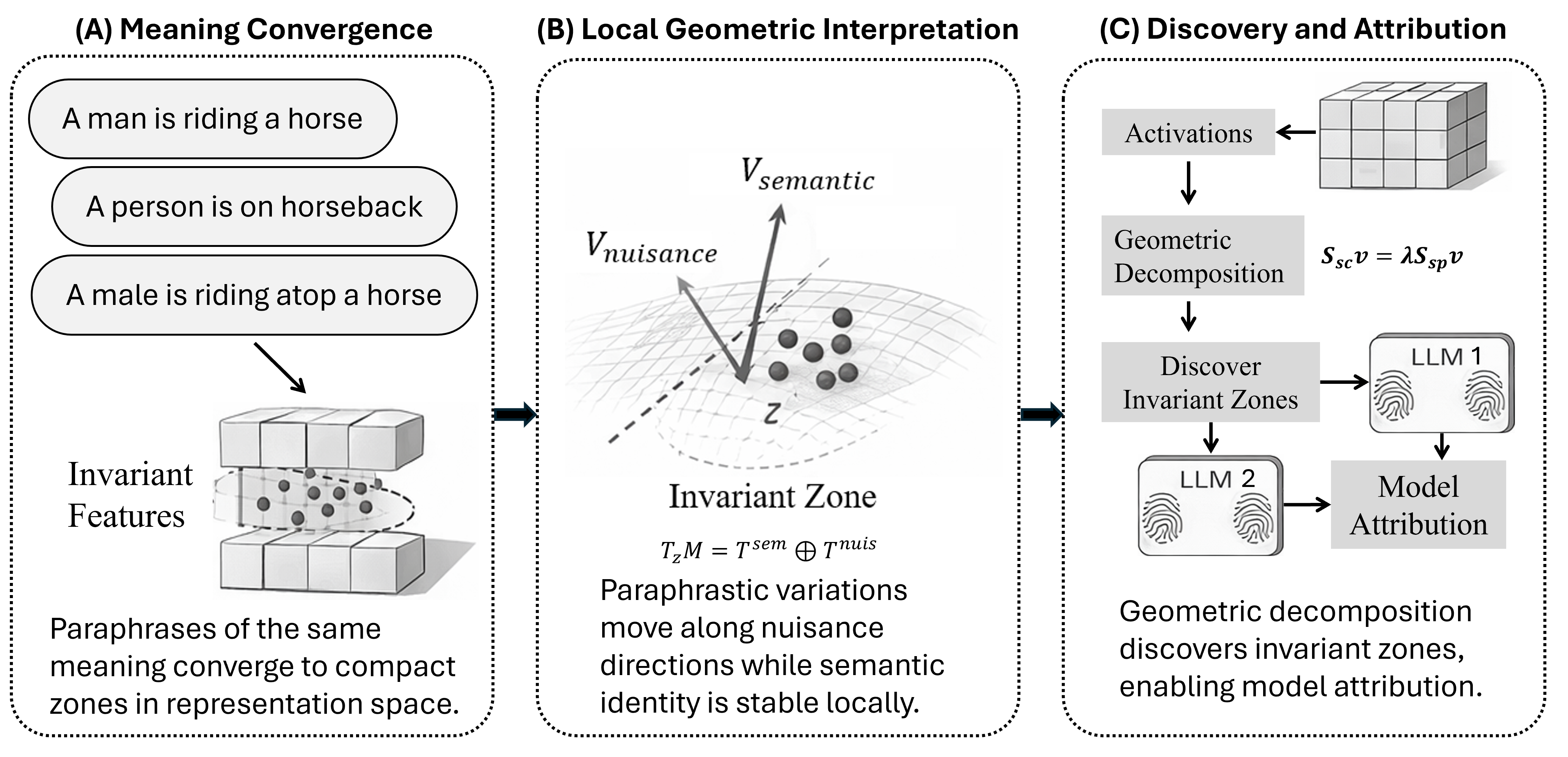}
    \vspace{-0.75em}
    \caption{Emergence of invariant semantic regions in LLMs. (A) Paraphrases propagate through layers and tend to converge to compact zones.
    (B) Local geometric interpretation: nuisance vs semantic tangent directions.
    (C) Geometric decomposition detects invariant zones for model attribution.}
    \label{fig:FIA}
    \vspace{-1.5em}
\end{figure}

Understanding invariant internal representations is important for both scientific and practical reasons. From a representation learning perspective, identifying where semantic stability arises provides insight into how transformers organize meaning across depth and how abstraction emerges during inference. From a systems and security perspective, invariant representations offer a mechanism to analyze model behavior beyond observable outputs. Because these internal structures reflect how a model organizes semantic information, they may support applications such as model attribution, forensic analysis, and robustness evaluation.

Existing efforts to analyze or identify language models span several complementary directions. Prior work on semantic representations and probing has shown that transformer activations encode structured linguistic and semantic information and exhibit robustness to paraphrasing. A separate line of work studies representation geometry through similarity metrics and subspace analysis, revealing that learned features admit meaningful low-dimensional structure across layers and models. More recently, model attribution and fingerprinting approaches have demonstrated that internal activations and outputs can contain model-specific signals. 
While these studies provide important insights, they largely treat semantic stability, representation geometry, and model identity as separate phenomena. In particular, existing approaches do not provide a local geometric characterization of semantic invariance as a structural property of latent representations, nor a principled mechanism to identify invariant structure by explicitly separating semantic-preserving and semantic-changing variation. As a result, it remains difficult to systematically understand how semantic invariance emerges across layers and how it relates to model-specific representation geometry.

In this work, we develop a geometric framework for analyzing semantic invariance in language models. As illustrated in Fig.~\ref{fig:FIA}, we hypothesize that semantically equivalent inputs form locally stable regions in latent space, where semantic-preserving and semantic-changing variations correspond to approximately distinct nuisance and semantic directions. 
(1) Building on this hypothesis, our first contribution is a local geometric definition of invariant latent features as representations that are stable under semantic-preserving transformations while remaining discriminative across meanings. 
(2) Our second contribution is a data-driven framework for identifying local invariant features via contrastive sensitivity decomposition, which separates semantic-changing from semantic-preserving variation.
(3) Our third contribution is to show that these invariant zones exhibit model-specific geometry and can serve as latent signatures for zero-shot model attribution, demonstrating that invariant representations encode both semantic stability and model individuality.

%% file: sections/related.tex
\textbf{Semantic Representations and Emergent Invariance in Language Models.}
Prior work shows that transformer-based language models develop structured representations across depth, with lower layers encoding lexical and syntactic information and higher layers capturing contextual and semantic properties~\cite{ethayarajh-2019-contextual,tenney-etal-2019-bert}. Probing studies further demonstrate that linguistic structures such as syntax and dependency relations can be recovered from these representations~\cite{hewitt-manning-2019,rogers-etal-2020}. Geometric analyses of BERT additionally suggest that semantic and syntactic information occupy structured subspaces in latent space, with word senses exhibiting fine-grained organization~\cite{reif2019visualizing}. Together, these findings indicate that language models encode semantic structure beyond surface form. 
This line of work also implies an empirical form of semantic invariance: inputs that differ in wording but preserve meaning often yield similar representations. Analyses of contextualized embeddings and their reductions to static representations further show that semantic information remains stable across contexts~\cite{ethayarajh-2019-contextual,bommasani-etal-2020-interpreting}.

Existing studies primarily establish that semantic structure is present in language-model representations. However, a local geometric characterization of how semantic invariance emerges across layers, how paraphrastic variation separates from semantic variation, and how such invariant structure can be operationally identified, remains less explored.

\textbf{Representation Geometry and Subspace Analysis in Neural Networks.}
A central line of work studies neural representations through geometric and statistical comparisons of activation spaces. Representational similarity analysis (RSA) and related methods quantify relationships across inputs, layers, and models. Widely used techniques include canonical correlation analysis (CCA) and its variants such as singular vector CCA (SVCCA)~\cite{raghu2017svcca} and projection-weighted CCA (PWCCA)~\cite{morcos2018pwcca}, as well as centered kernel alignment (CKA), which is invariant to orthogonal transformations and isotropic scaling~\cite{kornblith2019similarity}. These methods enable systematic comparison of representations and are widely used to analyze training dynamics, layer correspondence, and cross-model similarity. 
Complementary approaches examine the structure of representation spaces directly. Linear probes show that linguistic and semantic features can be recovered from learned representations, while geometric analyses indicate that such information resides in structured subspaces~\cite{hewitt-manning-2019,reif2019visualizing}. Together, these results suggest that neural representations admit meaningful low-dimensional structure that can be characterized through linear projections and subspace decompositions.

Despite these advances, existing approaches mainly measure similarity or identify directions associated with specific properties. They do not explicitly provide mechanisms for separating semantic-preserving from semantic-changing variation. Consequently, while prior work reveals structure in representation space, a principled framework for isolating invariant subspaces that distinguish semantic content from nuisance variation remains limited. This motivates our formulation of invariant subspace discovery as a contrastive geometric decomposition, where invariant directions arise from a generalized eigenvalue operator that separates semantic and nuisance components.

\textbf{Model-Specific Representations and Attribution via Internal Activations.}
A growing body of work suggests that language models encode model-specific structure despite similar external behavior. Evidence arises from transfer studies, where soft prompts optimized for one model often fail to generalize to others, indicating interaction with model-specific latent pathways~\cite{lester-etal-2021-power}. Related observations come from adversarial trigger analysis, where input-agnostic triggers reveal global biases and model-dependent vulnerabilities~\cite{wallace-etal-2019-universal}. These results suggest that similar linguistic functions may be implemented through different internal geometries. 
Model attribution has also been studied via generated text, watermarking, and explicit identification signals. Prior work explores source-model identification from outputs~\cite{antoun-etal-2024-text}, watermark-based attribution~\cite{lu-etal-2024-source, dasgupta2025watermarking}, and instruction-based identification through lightweight training interventions~\cite{xu-etal-2024-instructional}.

While these approaches demonstrate that model identity can be detected, they typically rely on output-level artifacts, task-specific classifiers, watermarking mechanisms, or training-induced signatures. In contrast, our work uses invariant latent structure itself as a compact semantic signature. Attribution is therefore not treated as a standalone identification mechanism, but as a downstream validation that invariant subspaces preserve both semantic stability and model-specific representation geometry.

%% file: sections/method.tex
We present a geometric framework for semantic invariance in language models, based on the hypothesis that paraphrastic variation induces structured zones of stability. Section~\ref{sec: define} formalizes invariant latent features, Section~\ref{sec: find} introduces a contrastive subspace discovery method, and Section~\ref{sec: attribution_app} uses the resulting representations for zero-shot model attribution as empirical validation.

\subsection{Invariant Latent Features: Definition and Local Geometry}
\label{sec: define}

Let \(f_m\) denote a language model, \(h_\ell(x)\in\mathbb{R}^{T\times d}\) denote the token-level hidden states at layer \(\ell\), and \(z=\Phi_\ell(x)\in\mathbb{R}^d\) denote their pooled representation, which serves as the primary object of analysis. 
We study invariance of these representations under semantic-preserving transformations. Formally, for a transformation \(T(\cdot)\) such as paraphrasing, invariance requires
\(
\Phi_\ell(T(x)) \approx \Phi_\ell(x).
\)
To avoid trivial collapse, representations must also remain discriminative across meanings. Specifically, for paraphrases \(x_1,x_2\) and an input \(x_3\) with different meaning, we require
\(
\Phi_\ell(x_1) \approx \Phi_\ell(x_2), \quad
\Phi_\ell(x_1) \not\approx \Phi_\ell(x_3).
\)
This within--between criterion defines invariant latent features as representations that are stable under semantic-preserving variation while remaining separable across semantic content. 

To interpret this property, we adopt a geometric perspective inspired by the manifold hypothesis~\cite{fefferman2016testing}. Although latent representations reside in a high-dimensional ambient space \(\mathbb{R}^d\), meaningful semantic information occupies only a structured subset of this space. Let \(\mathcal{M} \subset \mathbb{R}^d\) denote the semantic representation manifold. Inputs with coherent semantic meaning are mapped to points lying near this manifold, while arbitrary points in the ambient space generally do not correspond to meaningful language representations. 
Under this view, semantically equivalent inputs correspond to nearby points on the manifold. Paraphrasing can therefore be interpreted as a local perturbation that moves the representation within a neighborhood of the same semantic region. Specifically, for paraphrases \(x_1\) and \(x_2\),
\(
\|\Phi_\ell(x_1) - \Phi_\ell(x_2)\| \leq \epsilon,
\)
for some small \(\epsilon > 0\), while input \(x_3\) with distinct semantic meaning occupy separate regions of the manifold.

\begin{wrapfigure}{r}{0.70\textwidth}
  \centering
  \vspace{-1.0em}
  \includegraphics[width=1.0\linewidth]{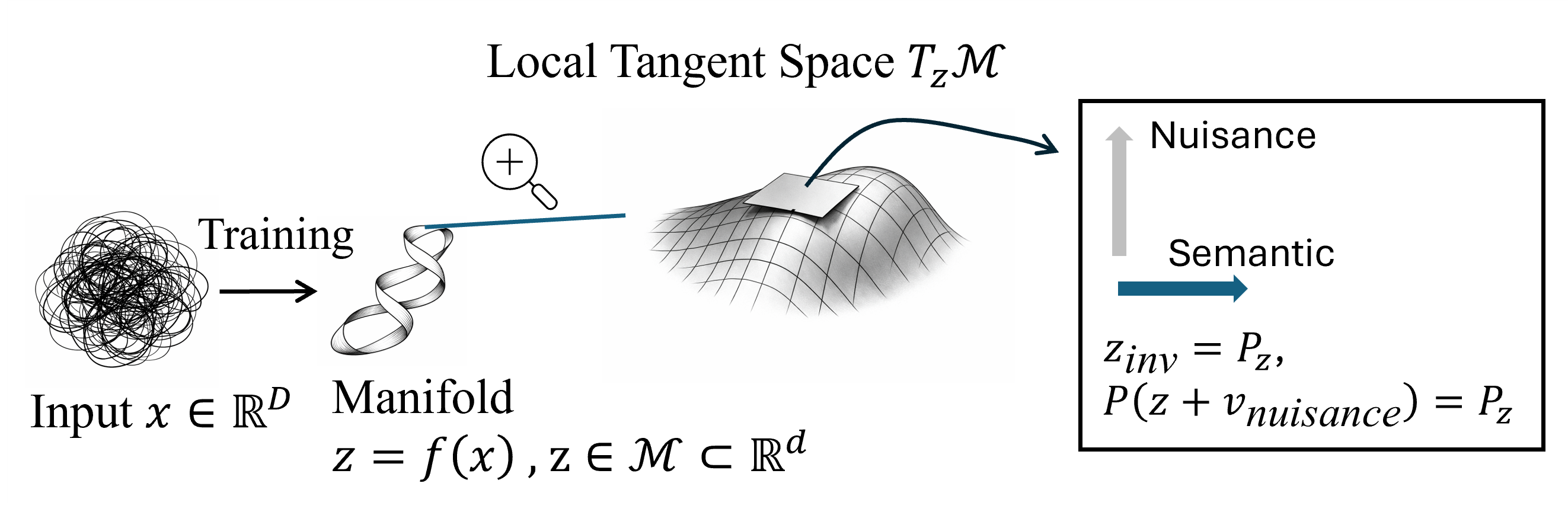}
  \vspace{-2.0em}
  \caption{
    Local geometric hypothesis of semantic invariance in LM representations. 
    During training, inputs $x \in \mathbb{R}^D$ are mapped by the model to latent representations 
    $z = f(x)$ that lie on a lower-dimensional manifold $\mathcal{M} \subset \mathbb{R}^d$. 
    At a point $z \in \mathcal{M}$, the tangent space $T_z\mathcal{M}$ provides a local linear 
    approximation of allowable variations in representation space. 
    Semantic-preserving transformations such as paraphrasing primarily induce movements along 
    nuisance directions that do not alter the underlying meaning. 
    Invariant latent features correspond to projections that remove these nuisance directions, 
    so that $z_{\text{inv}} = P_z$ and $P(z + v_{\text{nuisance}}) = P_z$, preserving semantic identity 
    while ignoring surface variation.
    }
  \label{fig:define}
  \vspace{-1.0em}
\end{wrapfigure}

We propose to analyze the local structure of \(\mathcal{M}\) through its tangent space. At a point \(z=\Phi_\ell(x)\) on the manifold, the tangent space \(T_z\mathcal{M}\) contains all directions along which the representation can move infinitesimally while remaining on the manifold. In other words, \(T_z\mathcal{M}\) provides a local linear approximation of the manifold and captures the allowable directions of variation in representation space. 
Figure~\ref{fig:define} illustrates this local geometric interpretation. Variations of representations on the manifold arise from two qualitatively different sources: semantic changes, which alter the underlying meaning of the input, and nuisance changes, which modify surface form while preserving semantic content. 
Locally, these two types of variation can be interpreted as approximately distinct directions within the tangent space. We therefore consider a working decomposition of the form
\begin{equation}
T_z\mathcal{M} \approx T_z^{\mathrm{semantic}} \cup T_z^{\mathrm{nuisance}}
\label{eq: tan_decompose}
\end{equation}
where $T^{\text{semantic}}_z$ captures directions associated with changes in semantic content, and $T^{\text{nuisance}}_z$ captures directions corresponding to semantic-preserving variation. In practice, this decomposition is a local and approximate factorization, since learned representations may exhibit partial overlap and entanglement between these components. 
Under a semantic-preserving transformation such as paraphrasing, the representation primarily moves along nuisance directions,
\(
z' = z + v_{\text{nuisance}} .
\)
Invariant features correspond to representations that ignore such nuisance motion. If \(P\) denotes a projection operator onto the semantic subspace, the invariant representation can be written as
\(
z_{\text{inv}} = Pz,
\)
which satisfies
\(
P(z + v_{\text{nuisance}}) = Pz.
\)
Throughout this work, ``invariant'' refers specifically to invariance with respect to semantic-preserving transformations. Thus, invariant features are expected to remain stable under paraphrasing while remaining sensitive to semantic-changing perturbations.

This local geometric interpretation provides an explanation for why invariant latent features naturally emerge during training. To maintain consistent behavior under paraphrasing, the model must map semantically equivalent inputs to nearby regions in latent space while preserving separability across meanings. These constraints induce structured zones of stability on the representation manifold where paraphrase variants cluster while remaining distinguishable from other semantic concepts.

Although the existence of such invariant zones follows from the need for semantic consistency, their geometric realization is generally model-specific. Transformer training involves highly non-convex optimization~\cite{kawaguchi2016deep, choromanska2015loss}, and different models may converge to distinct embeddings of the underlying semantic manifold due to variations in initialization, architecture, training data, and optimization dynamics. Consequently, even when two models encode similar semantic concepts, the corresponding latent manifolds may differ in orientation, curvature, and local geometry. Invariant latent features therefore reflect how a particular model internally organizes semantic information. This combination of semantic stability and geometric individuality later enables these invariant representations to serve as distinctive identifiers for model attribution.

\subsection{Invariant Subspace Identification via Contrastive Sensitivity Decomposition}
\label{sec: find}

The geometric formulation in Section~\ref{sec: define} characterizes invariant latent features as directions that remain stable under semantic-preserving transformations while remaining responsive to semantic change. We now introduce a data-driven procedure to identify such directions from model activations. The key idea is to contrast two sources of variation in latent space: variation induced by semantic-preserving transformations, which primarily reflects nuisance motion, and variation induced by semantic-changing transformations, which reflects movement across meanings.

Let \(x\) denote an input prompt, at layer $\ell$, let $h_\ell(x) \in \mathbb{R}^{T \times d}$ denote token-level hidden states. We define a representation vector $z_\ell(x)=\Phi_\ell(x)\in\mathbb{R}^d$ as a deterministic pooling of $h_\ell(x)$. 
In practice, we use the final-token hidden state, i.e., $z_\ell(x)=h_\ell^{(T)}(x)$, which incorporates information from preceding tokens and directly supports next-token prediction. We compare this choice with mean pooling in Appendix~\ref{app:pooling}.
For each input, we construct two perturbation families. The first consists of semantic-preserving variants, denoted by \(\{x_i^{\mathrm{sp}}\}_{i=1}^{n}\), such as paraphrases that preserve the underlying meaning while altering surface form. The second consists of semantic-changing variants, denoted by \(\{x_j^{\mathrm{sc}}\}_{j=1}^{m}\), whose semantic content differs from that of the anchor input. We then compute the corresponding finite-difference feature changes
\(
\Delta z_i^{\mathrm{sp}}=\Phi_\ell(x_i^{\mathrm{sp}})-\Phi_\ell(x),
\)
\(
\Delta z_j^{\mathrm{sc}}=\Phi_\ell(x_j^{\mathrm{sc}})-\Phi_\ell(x).
\)
These difference vectors provide empirical estimates of local motion in the representation space. Under the local geometric hypothesis, semantic-preserving perturbations primarily trace nuisance directions in the tangent space, whereas semantic-changing perturbations additionally traverse semantic directions. To capture these two modes of variation, we define the empirical sensitivity covariances
\(
S_{\ell}^{\mathrm{sp}}=
\frac{1}{n}\sum_{i=1}^{n}\Delta z_i^{\mathrm{sp}}(\Delta z_i^{\mathrm{sp}})^\top,
\)
\(
S_{\ell}^{\mathrm{sc}}=
\frac{1}{m}\sum_{j=1}^{m}\Delta z_j^{\mathrm{sc}}(\Delta z_j^{\mathrm{sc}})^\top.
\) 
We then identify contrastive directions by solving the generalized eigenvalue problem:
\begin{equation}
   S_{\ell}^{\mathrm{sc}} v = \lambda \, S_{\ell}^{\mathrm{sp}} v. 
\label{eq: cross_eigen}
\end{equation}
Each eigenvector \(v\) maximizes the Rayleigh quotient
\(
\lambda(v)=
v^\top S_{\ell}^{\mathrm{sc}} v / v^\top S_{\ell}^{\mathrm{sp}} v.
\)
This formulation encodes the desired notion of the proposed invariance. 
Directions with large \(\lambda\) exhibit strong variation under semantic change but weak variation under semantic-preserving perturbations, and therefore correspond to invariant semantic directions. In contrast, directions with small \(\lambda\) are dominated by semantic-preserving variation and are interpreted as nuisance directions. 
We therefore use the term invariant direction to denote a direction that is stable under semantic-preserving perturbations but discriminative under semantic-changing perturbations. 
All covariance estimates and the generalized eigenvalue problem are computed on $z_\ell(x)$.

In practice, $S_{\mathrm{sp}}$ is low rank because the number of semantic-preserving perturbations is much smaller than the hidden dimension. We therefore solve a regularized generalized eigenvalue problem, which ensures positive definiteness of the denominator operator and yields a well-posed decomposition,
\(
S_{\mathrm{sc}} v = \lambda (S_{\mathrm{sp}} + \epsilon I)v,
\)
where $\epsilon > 0$ ensures numerical stability. Let $V_{\ell}^{\mathrm{inv}} = [v_1, \ldots, v_k]$ denote the top-$k$ generalized eigenvectors with the largest eigenvalues at layer $\ell$. 
Since generalized eigenvectors are orthogonal with respect to the $(S_{\mathrm{sp}} + \epsilon I)$-inner product but not necessarily under the Euclidean inner product, we compute an orthonormal basis
\(
U_\ell^{\mathrm{inv}} = \mathrm{orth}(V_\ell^{\mathrm{inv}})
\)
via QR decomposition. This ensures that $U_\ell^{\mathrm{inv}} U_\ell^{\mathrm{inv}\top}$ defines a valid Euclidean projection operator. 
The invariant representation at layer $\ell$ is then obtained by projection,
\begin{equation}
z_{\mathrm{inv}}
=
U_\ell^{\mathrm{inv}}
U_\ell^{\mathrm{inv}\top} z .
\label{eq: decompose_z_inv}
\end{equation}
This projection suppresses nuisance-sensitive components while preserving directions that remain stable under paraphrasing yet discriminate across meanings. 
The complementary nuisance component is given by
\(
z_{\mathrm{nuis}}
=
\left(
I-
U_\ell^{\mathrm{inv}}
U_\ell^{\mathrm{inv}\top}
\right) z.
\)
Together they induce the decomposition
\(
z = z_{\mathrm{inv}} + z_{\mathrm{nuis}},
\)
providing a constructive approximation of the semantic--nuisance decomposition in Eq.~\ref{eq: tan_decompose}. In particular, \(z_{\mathrm{inv}}\) captures components aligned with invariant semantic directions, while \(z_{\mathrm{nuis}}\) captures variation dominated by nuisance-sensitive directions.

The decomposition in Eq.~\ref{eq: decompose_z_inv} also admits a local differential interpretation. When perturbations are infinitesimal, \(\Delta z \approx J_{\Phi_\ell}(x)\Delta x\), where \(J_{\Phi_\ell}(x)\) is the Jacobian of the layer-wise representation map. The resulting covariance operators satisfy
\(
S \approx J_{\Phi_\ell}(x)\,\mathbb{E}[\Delta x \Delta x^\top]\,J_{\Phi_\ell}(x)^\top.
\)
Hence, the proposed covariance-based formulation can be viewed as a data-driven approximation of Jacobian sensitivity. Moreover, if the input perturbations are isotropic so that \(\mathbb{E}[\Delta x \Delta x^\top]=I\), then
\(
S \approx U \Sigma^2 U^\top,
\)
where \(J_{\Phi_\ell}(x)=U\Sigma V^\top\) is the singular value decomposition of the Jacobian. In that case, the eigenvectors of \(S\) align with the left singular vectors of the Jacobian, and the eigenvalues correspond to squared singular values. The generalized eigenvalue formulation therefore extends standard sensitivity analysis by explicitly separating semantic and nuisance variation.

We apply this procedure across layers and select those exhibiting strong invariant structure based on their generalized eigenvalue spectra. Invariant zones are thus defined as sets of layers associated with leading eigenvalues, without requiring spatial contiguity. 
For each model \(m\) and input \(x\), we define the invariant zone representation by aggregating invariant projections across the selected layers,
\(
s_m(x)
=
\mathrm{Agg}
\big(
z_{\mathrm{inv}}^{(\ell_i)},
\ldots,
z_{\mathrm{inv}}^{(\ell_j)}
\big),
\)
where \(\{\ell_i,\ldots,\ell_j\}\) denotes the set of layers whose invariant subspaces are identified by large generalized eigenvalues, and \(\mathrm{Agg}(\cdot)\) denotes feature aggregation across layers, implemented as concatenation of the invariant representations.
This representation serves as a compact semantic signature capturing the invariant structure of the model.

\subsection{Zero-shot Model Attribution Using Invariant Features}
\label{sec: attribution_app}

\begin{wrapfigure}{r}{0.35\textwidth}
  \centering
  \vspace{-1.5em}
  \includegraphics[width=1.0\linewidth]{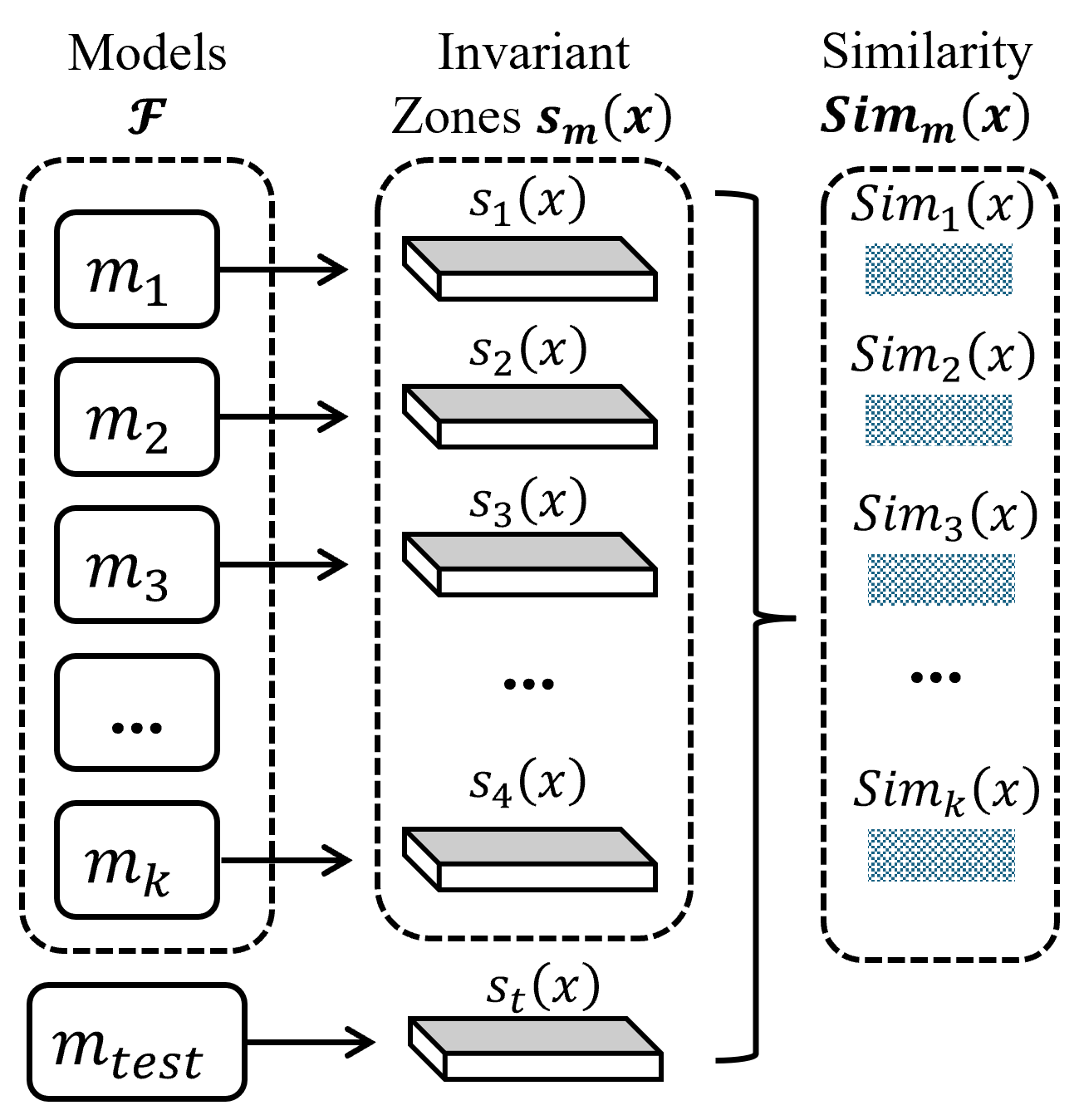}
  \vspace{-2.0em}
    \caption{
    Zero-shot model attribution via invariant zone signatures. 
    Each model \(m\) is represented by an invariant signature \(s_m(x)\). 
    Given a test model, attribution selects the reference model with highest similarity in invariant space.
    }
    \label{fig:attribution}
  \vspace{-1.5em}
\end{wrapfigure}

The preceding section constructs invariant zones and their corresponding representations \(s_m(x)\), which capture model-specific semantic structure while remaining stable under paraphrasing. We now use these representations for model attribution.

\textbf{Attribution Formulation.}
Let
\(
\mathcal{F}=\{m_1,\ldots,m_K\}
\)
denote a collection of language models. Given an input prompt \(x\), the attribution task is to infer its originating model \(m\in\mathcal{F}\) using invariant zone representations. For each reference model \(m\), we compute its signature \(s_m(x)\), and for a test model we compute \(s_t(x)\). Attribution is then performed by similarity matching
\(
\hat{m}
=
\arg\max_{m\in\mathcal{F}}
\;
\mathrm{Sim}\big(s_t(x),\, s_m(x)\big),
\)
where \(\mathrm{Sim}(\cdot,\cdot)\) denotes a similarity measure such as cosine similarity. 
This formulation corresponds to a zero-shot prototypical~\cite{snell2017prototypical} style classification scheme in invariant representation space, as illustrated in Fig.~\ref{fig:attribution}. Each model is represented by its invariant zone, and attribution is performed by selecting the closest match.

\textbf{Interpretation as Hypothesis Validation.}
Attribution is not posed as a standalone forensic task, but as validation of the proposed geometric framework. If invariant zone representations retain discriminative model identity, this indicates that the discovered invariant structures encode model-specific organization of semantic information. In this way, attribution provides empirical evidence that semantic invariance in language models is a geometric property of latent representations.

%% file: sections/experiment.tex
\begin{wraptable}{r}{0.29\textwidth}
\vspace{-1.0em}
\centering
\footnotesize
\setlength{\tabcolsep}{5pt}
\renewcommand{\arraystretch}{1.1}
\begin{tabular}{lr}
\toprule
\textbf{Component} & \textbf{Count} \\
\midrule
Semantic groups & 1,400 \\
Domains & 13 \\
\midrule
Anchors/group & 1 \\
SP/group & 20 \\
Soft SC/group & 10 \\
Hard SC/group & 10 \\
\midrule
Total sentences & 57,400 \\
\bottomrule
\end{tabular}
\caption{Dataset composition. Each of the 1,400 semantic groups contains one anchor, 20 SP, 10 soft SC, and 10 hard SC variants.}
\label{tab:dataset_category}
\vspace{-4.0em}
\end{wraptable}

We evaluate the proposed geometric framework through a series of analyses that examine the emergence, structure, and functional role of invariant features across models and layers. Core experiments focus on dataset and model setup, invariant zone localization, geometric validation, causal intervention, and attribution, while extended ablations and additional analyses are deferred to the appendix for completeness. 

\subsection{Evaluation Setup}
\label{sec:eval_setup}

\textbf{Dataset Construction.}
We construct a dataset of 1,400 semantic groups comprising 57,400 sentences across 13 domains (Table~\ref{tab:dataset_category}), covering diverse topics and linguistic contexts to investigate semantic invariance under controlled SP and SC perturbations.
Each group defines a local semantic neighborhood centered on an anchor input. SP variants preserve the underlying meaning while varying lexical form, grammatical structure, and writing style, including formal/casual and question/statement reformulations. Soft SC variants modify an attribute or semantic slot while retaining the original topic, whereas hard SC variants introduce domain- or topic-level semantic changes. For example, given the anchor ``What is the official language of France?'', an SP variant is ``Which language holds official status in France?'', a soft SC variant is ``What is the national currency of France?'', and a hard SC variant is ``What is the longest river in South America?''. Each SC example includes edit-type and source/target-domain metadata, enabling perturbation-specific analysis. Additional construction details and examples are provided in Appendix~\ref{app:expanded_data}.

\begin{wraptable}{r}{0.375\textwidth}
\vspace{-1.0em}
\centering
\footnotesize
\setlength{\tabcolsep}{4pt}
\renewcommand{\arraystretch}{1.1}
\begin{tabular}{llc}
\toprule
\textbf{Model} & \textbf{Family} & \textbf{Params} \\
\midrule
Mistral-Instruct-v0.3 & Mistral  & 7B \\
Llama-3-Instruct      & Llama    & 8B \\
Gemma-IT              & Gemma    & 7B \\
Qwen2.5-Instruct      & Qwen     & 7B \\
GLM-4-Chat            & GLM      & 9B \\
DeepSeek-MoE-Chat     & DeepSeek & 16B \\
InternLM2-Chat        & InternLM & 7B \\
Phi-4-Mini            & Phi      & 4B \\
Falcon3-Instruct      & Falcon   & 7B \\
\bottomrule
\end{tabular}
\caption{Language models used for evaluation, spanning diverse families, scales, and training paradigms.}
\label{tab:models}
\vspace{-1.5em}
\end{wraptable}

To prevent data leakage, each of the 1,400 semantic groups is divided into disjoint discovery and validation subsets, each containing 10 SP and 10 SC variants (5 soft and 5 hard). The discovery subset is used exclusively for invariant subspace identification and layer selection through generalized eigenvalue analysis, while the held-out validation subset is used for downstream evaluation, including tangent-direction validation, representation intervention, projection analysis, and model attribution. This design preserves the local nature of our framework: each group defines an anchor-centered semantic neighborhood with its own invariant subspace. The consistent results across groups support the existence of repeatable local geometric structure rather than a universal invariant projection.

\textbf{Models Analyzed.}
We evaluate nine transformer-based language models listed in Table~\ref{tab:models}, spanning diverse architectural families, parameter scales (4B–16B), and training paradigms. The suite includes models trained under different objectives and pipelines, including instruction tuning, chat alignment, and mixture-of-experts architectures. This diversity allows us to assess whether invariant structure is a consistent property of representation geometry across model families, rather than an artifact of a specific architecture, scale, or training procedure.

\subsection{Localization of Invariant Feature Zones}
\label{sec: localize}

\begin{wrapfigure}{r}{0.4\textwidth}
    \vspace{-1.0em}
    \centering
    \includegraphics[width=1.0\linewidth]{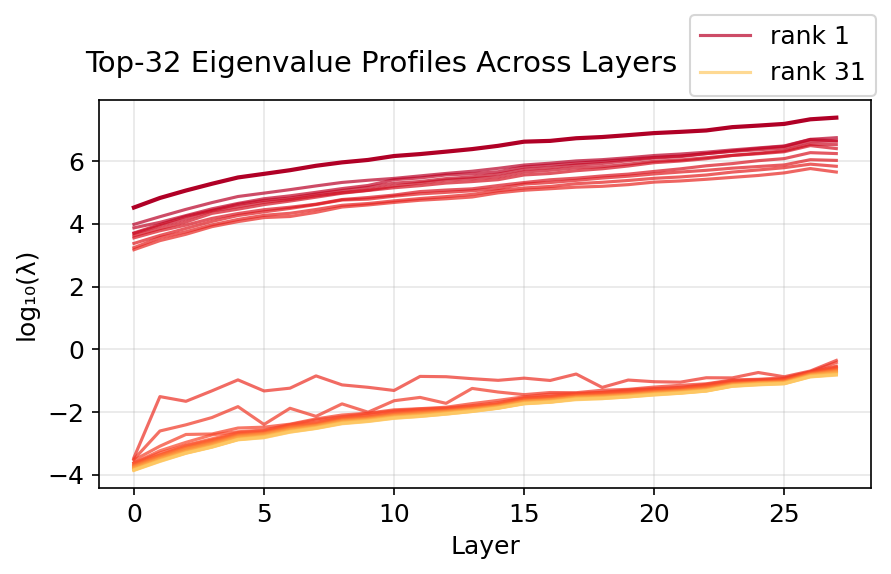}
    \vspace{-2.0em}
    \caption{Per-layer eigenvalue profiles for $k=32$ invariant directions (log scale). Each line denotes an eigenvector rank. Eigenvalues increase with depth, and higher-ranked directions remain dominant, indicating a stable and ordered decomposition of invariant signal across layers.}
    \label{fig:eigenvalue_profiles}
    \vspace{-1.0em}
\end{wrapfigure}

We investigate where invariant structure emerges across model depth by analyzing the layer-wise generalized eigenvalue spectrum defined in Section~\ref{sec: find}. The goal is to identify layers in which semantic-changing variation is maximally separated from semantic-preserving variation, corresponding to the emergence of invariant feature zones. 
Since invariant zones are locally defined for each semantic group and model pair, the following analysis presents one representative example from Llama, illustrating how such zones emerge across depth within a specific local semantic neighborhood. 
For each layer $\ell$, we compute the generalized eigenvalues $\{\lambda_i^{(\ell)}\}_{i=1}^k$ from Eq.~\ref{eq: cross_eigen}, which quantify the ratio between semantic-changing and semantic-preserving variation along each direction. Large eigenvalues indicate directions that are stable under paraphrasing but responsive to semantic change, and therefore correspond to invariant features. We analyze the distribution of the top-$k$ eigenvalues across layers to characterize how invariant structure evolves with depth.

\textbf{Invariant structure becomes more pronounced in mid-to-late layers.} 
Figure~\ref{fig:eigenvalue_profiles} shows the per-layer eigenvalue profiles on a log scale. The leading eigenvalues generally increase with depth, indicating stronger separation between semantic-changing and semantic-preserving variation. However, invariant zones are selected using eigenvalues normalized by the total positive eigenvalue mass within each layer, rather than raw magnitudes alone, reducing bias toward deeper layers. The relative magnitudes of the leading eigenvalues reveal a small set of dominant semantic-sensitive directions, with the strongest separation emerging in mid-to-late layers. This observation is consistent with the local geometric hypothesis in Section~\ref{sec: define}, suggesting that semantic invariance is concentrated in specific depth regions rather than uniformly distributed across layers. These layers are subsequently used to define invariant zones.

\subsection{Tangent-Direction Validation of Invariant Geometry}
\label{sec: tangetD}

We empirically validate the geometric decomposition introduced in Section~\ref{sec: find} by testing whether paraphrase-induced variation concentrates in the nuisance subspace, while semantic displacement lies predominantly outside it. This experiment directly evaluates the geometric structure discovered via contrastive sensitivity decomposition. 

\begin{wraptable}{r}{0.35\textwidth}
\vspace{-1.0em}
\centering
\footnotesize
\setlength{\tabcolsep}{4pt}
\renewcommand{\arraystretch}{1.2}
\begin{tabular}{c|ccc}
\hline
Model 
& G1 
& G4 
& G6 \\
\hline

Mistral 
& 0.0189 & 0.0213 & 0.0198 \\

Llama3 
& 0.0127 & 0.0141 & 0.0135 \\

Qwen2.5
& 0.0158 & 0.0172 & 0.0164 \\

Gemma
& 0.0169 & 0.0184 & 0.0176 \\

Phi4-mini
& 0.0215 & 0.0231 & 0.0223 \\

GLM4
& 0.0112 & 0.0126 & 0.0119 \\

DeepSeek
& 0.0105 & 0.0118 & 0.0112 \\

InternLM2
& 0.0121 & 0.0134 & 0.0128 \\

Falcon3
& 0.0116 & 0.0129 & 0.0123 \\

\hline
\end{tabular}
\caption{Nuisance projection energy $E_{\mathrm{nuis}}$ across models and groups. Low values indicate that semantic displacement lies largely outside the dominant nuisance-sensitive directions.}
\label{tab:energy}
\vspace{-2.0em}
\end{wraptable}

Let the nuisance component at layer $\ell$ be defined as the Euclidean orthogonal complement of the invariant subspace. Given $U_\ell^{\mathrm{inv}} \in \mathbb{R}^{d \times k}$, the nuisance component of a representation is obtained via the projection operator $(I - U_\ell^{\mathrm{inv}} U_\ell^{\mathrm{inv}\top})$. We do not explicitly construct a nuisance basis, as lower-ranked generalized eigenvectors are not guaranteed to form a stable or Euclidean-orthogonal complement in high-dimensional settings. 
For each semantic group $g=\{x_1,\ldots,x_{n_g}\}$, we compute latent representations $z_i=\Phi_\ell(x_i)$ and define the group centroid $\bar{z}_g=\frac{1}{n_g}\sum_{i=1}^{n_g}z_i$. 
For two semantic groups $g$ and $h$, we define the semantic displacement
\(
\Delta_{gh} = \bar z_g - \bar z_h.
\)
We then evaluate how much of this displacement lies within the nuisance subspace using the projection energy
\(
E_{\mathrm{nuis}} =
\left\|(I - U_\ell^{\mathrm{inv}} U_\ell^{\mathrm{inv}\top}) \, \Delta_{gh}\right\|_2^2  / 
\|\Delta_{gh}\|_2^2.
\) 
Under the local geometric hypothesis, semantic-preserving variations span the nuisance subspace, while semantic differences lie in complementary directions. In this case, the projection operator $(I-U^{\mathrm{inv}}_\ell U^{\mathrm{inv}\top}_\ell)$ removes the semantic component of $\Delta_{gh}$, leaving little residual energy. Consequently, the projected vector $(I-U^{\mathrm{inv}}_\ell U^{\mathrm{inv}\top}_\ell)\Delta_{gh}$ has small norm, leading to $E_{\mathrm{nuis}} \approx 0$. 
Geometrically, $E_{\mathrm{nuis}}$ measures the fraction of semantic displacement energy aligned with nuisance directions. When this value approaches zero, it implies that $\Delta_{gh}$ is nearly orthogonal to the nuisance subspace, and therefore lies primarily in the complementary semantic subspace. This behavior directly reflects the tangent-space decomposition in Eq.~\ref{eq: tan_decompose}, where semantic and nuisance variations occupy distinct, approximately orthogonal directions in the local representation geometry. 
Additional implementation details are provided in Appendix~\ref{appendix: implementation}.

\textbf{Semantic displacement lies predominantly outside the nuisance subspace.}
Table~\ref{tab:energy} reports $E_{\mathrm{nuis}}$ across multiple models and semantic groups. We consistently observe low nuisance projection energy ($\approx 0.01$--$0.02$), indicating that semantic displacement is concentrated in the invariant subspace. This pattern is consistent across models and groups, supporting the proposed local geometric separation between semantic and nuisance directions. Additional comparisons with PCA and cPCA baselines are provided in Appendix~\ref{app:projection_comparison}.

\subsection{Ablation Study: Causal Validation via Representation Intervention}
\label{sec:ab_mainpaper}

We evaluate whether the invariant subspace identified in Section~\ref{sec: find} causally encodes semantic content through representation-level interventions. Rather than relying solely on correlational analysis, these experiments directly perturb the invariant component of the representation and measure the resulting change in model outputs under both within-group and between-group conditions. 
Given the layer-wise decomposition
\(
z = z_{\mathrm{inv}} + z_{\mathrm{nuis}}
\)
obtained from Eq.~\ref{eq: decompose_z_inv}, we construct interventions by replacing the invariant component of an anchor representation while keeping the nuisance component fixed. In practice, interventions are implemented at the token level using the final-token hidden state $h_\ell^{(T)}(x)$, which directly determines next-token prediction in causal language models. We therefore set
\(
z_\ell(x) = h_\ell^{(T)}(x),
\)
so that
\(
h_\ell^{(T)}(x)=h_{\mathrm{inv}}+h_{\mathrm{nuis}}.
\) 
For an anchor hidden state $h^{(a)}$ and donor hidden state $h^{(d)}$, the intervened representation is
\(
h^{\mathrm{int}}
=
h^{(a)}
+
\alpha
\big(
h_{\mathrm{inv}}^{(d)}
-
h_{\mathrm{inv}}^{(a)}
\big),
\)
which is then propagated through the remaining layers. The intervention effect is measured using the KL divergence between the original and intervened next-token distributions,
\(
\mathrm{KL}(\ell)
=
D_{\mathrm{KL}}
\!\left(
p_{\mathrm{base}}
\,\|\,
p_{\mathrm{int}}
\right).
\) 

\begin{wraptable}{r}{0.45\textwidth}
\vspace{+0.25em}
\centering
\footnotesize
\setlength{\tabcolsep}{3pt}
\renewcommand{\arraystretch}{1.1}
\begin{tabular}{lccc}
\toprule
\textbf{Model} 
& \textbf{Between} 
& \textbf{Within-SP} 
& \textbf{Within-SC} \\
& KL$\uparrow$ & KL$\downarrow$ & KL$\uparrow$ \\
\midrule
Mistral-7B       & 0.974 & 0.036 & 0.341 \\
Llama-3-8B       & 0.986 & 0.040 & 0.360 \\
Gemma-7B         & 0.979 & 0.042 & 0.382 \\
Qwen2.5-7B       & 0.984 & 0.047 & 0.401 \\
GLM-4-9B         & 0.983 & 0.034 & 0.337 \\
DeepSeek-MoE     & 0.985 & 0.052 & 0.427 \\
InternLM2-7B     & 0.984 & 0.039 & 0.354 \\
Phi-4-mini       & 0.981 & 0.045 & 0.391 \\
Falcon3-7B       & 0.986 & 0.049 & 0.418 \\
\bottomrule
\end{tabular}
\caption{Invariant-component intervention results. Between-group swaps use donors from a different semantic group; within-group swaps use SP or SC donors from the same group. Low within-SP KL and high within-SC/between-group KL indicate semantic selectivity of the invariant subspace.}
\label{tab:inv_intervention_combined}
\vspace{-1.5em}
\end{wraptable}

We consider three intervention settings. In the within-group SP setting, the donor is a paraphrase from the same semantic group. In the within-group SC setting, the donor is semantically different but remains within the same semantic neighborhood. In the between-group setting, the donor is drawn from a completely different semantic group. Under the geometric hypothesis, SP interventions should produce minimal output change, while SC and between-group interventions should induce substantially larger shifts due to semantic inconsistency. 

\textbf{Invariant-component interventions produce substantially larger output changes under semantic-changing than semantic-preserving perturbations.}
Table~\ref{tab:inv_intervention_combined} summarizes KL divergence across all three intervention settings. Within-group SP interventions consistently produce near-zero KL divergence across models, indicating that the discovered invariant representation remains stable under paraphrastic variation. In contrast, within-group SC interventions yield substantially larger KL divergence, indicating that the invariant subspace is sensitive to semantic change even within a locally related semantic region. The effect becomes even stronger in the between-group setting, where invariant-component replacement produces consistently large KL divergence across all models. Together, these results provide evidence consistent with a causal role of the discovered invariant subspace in semantically meaningful model behavior. The separation between invariant- and nuisance-component interventions is further supported by narrow 95\% bootstrap confidence intervals across semantic groups (Appendix~\ref{app:bootstrap}) and remains evident in multi-hop reasoning tasks (Appendix~\ref{app:multihop}).

\subsection{Robust Model Attribution via Invariant Features}

\begin{wraptable}{r}{0.6\textwidth}
\vspace{-1.0em}
\centering
\scriptsize
\setlength{\tabcolsep}{4.5pt}
\renewcommand{\arraystretch}{1.3}
\begin{tabular}{c|ccc|ccc|ccc}
\hline
\multirow{2}{*}{Model} 
& \multicolumn{3}{c|}{Group1}
& \multicolumn{3}{c|}{Group2}
& \multicolumn{3}{c}{Group3} \\

& B & FT & D
& B & FT & D
& B & FT & D \\
\hline

Mistral 
& 93.2 & 91.4 & 89.8 
& 92.7 & 91.0 & 89.4 
& 93.5 & 91.8 & 90.2 \\

Llama3
& 94.1 & 92.5 & 90.9 
& 93.8 & 92.1 & 90.5 
& 94.4 & 92.9 & 91.3 \\

Gemma 
& 92.8 & 90.9 & 89.2 
& 92.3 & 90.5 & 88.8 
& 93.0 & 91.2 & 89.6 \\

Qwen2.5
& 93.6 & 91.8 & 90.1 
& 93.1 & 91.3 & 89.7 
& 93.9 & 92.0 & 90.5 \\

Phi4-mini
& 89.6 & 87.8 & 86.1 
& 89.2 & 87.4 & 85.8 
& 90.1 & 88.2 & 86.5 \\

GLM4
& 95.2 & 93.6 & 91.9
& 94.8 & 93.1 & 91.5
& 95.6 & 94.0 & 92.3 \\

DeepSeek
& 95.8 & 94.2 & 92.6
& 95.3 & 93.7 & 92.0
& 96.1 & 94.5 & 92.9 \\

InternLM2
& 94.9 & 93.3 & 91.7
& 94.5 & 92.9 & 91.3
& 95.2 & 93.6 & 91.9 \\

Falcon3
& 95.5 & 93.9 & 92.2
& 95.0 & 93.4 & 91.8
& 95.8 & 94.2 & 92.5 \\

\hline
\end{tabular}
\caption{Attribution accuracy (\%) across Base (B), Fine-tuned (FT), and Distilled (D) variants for three representative groups. Performance remains high and stable under model adaptation.}
\label{tab:model_wise_attribution}
\vspace{-1.25em}
\end{wraptable}

For each model $m \in \mathcal{F}$, attribution is performed by comparing the invariant representation of a test model to reference model signatures using the method defined in Section~\ref{sec: attribution_app}. 
To assess robustness, we consider three variants of each model: the original base pretrained model (B), a fine-tuned version (FT), and a distilled version (D). Fine-tuned models are obtained via instruction tuning on downstream tasks, while distilled models are trained to approximate teacher outputs with reduced capacity. Importantly, prototype invariant zones are computed from the original models and reused without modification for testing the FT and Dist variants. This protocol isolates whether invariant features reflect intrinsic representation geometry rather than training-specific adaptations. Additional implementation details for fine-tuned and distilled model variants, checkpoint construction, layer-selection protocol, and attribution evaluation are provided in Appendix~\ref{app:model_variants}.

\textbf{Invariant representations enable accurate model attribution and remain robust under fine-tuning and distillation.}
Table~\ref{tab:model_wise_attribution} reports attribution accuracy across models and variants. We observe consistently high accuracy for original base models (typically above $92\%$), indicating that invariant zone representations capture strong model-specific signatures. Attribution performance remains high under both fine-tuning and distillation, with only modest degradation (approximately $2$--$4\%$ across models). These results suggest that the discovered invariant features capture model-specific geometric structure that persists under model adaptation, despite changes in task behavior and model parameterization. 

\begin{wraptable}{r}{0.35\textwidth}
\vspace{+0.5em}
\centering
\scriptsize
\setlength{\tabcolsep}{5pt}
\renewcommand{\arraystretch}{1.2}
\begin{tabular}{lccc}
\toprule
\textbf{Variant} & \textbf{PCA} & \textbf{cPCA} & \textbf{Ours} \\
\midrule
Base & 86.08 & 89.33 & \textbf{95.45} \\
Fine-tuned & 83.98 & 87.53 & \textbf{94.20} \\
Distilled & 81.25 & 84.65 & \textbf{92.25} \\
\bottomrule
\end{tabular}
\caption{Zero-shot attribution accuracy (\%) averaged over four models (Llama, Mistral, Qwen2.5, and Gemma) with matched subspace dimensionality. Unlike Table~\ref{tab:model_wise_attribution}, these results are aggregated over the full evaluation set. Full results are in Appendix~\ref{app:attribution_baseline}.}
\label{tab:attribution_baseline_main}
\vspace{-1.25em}
\end{wraptable}

\textbf{Contrastive invariant decomposition outperforms variance-based and contrastive baselines in model attribution.}
To determine whether attribution performance arises from the proposed invariant geometry rather than generic representation structure, we compare our method against PCA and contrastive PCA (cPCA) using matched subspace dimensionality. PCA selects directions of dominant representation variance, while cPCA identifies directions maximizing the contrast $S_{\mathrm{SC}}-\alpha S_{\mathrm{SP}}$. In contrast, our generalized eigenvalue formulation identifies directions with high SC sensitivity relative to SP variation. Table~\ref{tab:attribution_baseline_main} summarizes accuracy averaged across four model families. Our method consistently outperforms both baselines under all adaptation settings, improving over cPCA by 6.13, 6.68, and 7.60 percentage points for base, fine-tuned, and distilled models, respectively. These results indicate that normalizing semantic-changing variation relative to semantic-preserving sensitivity provides more discriminative model signatures than variance maximization or additive covariance contrast alone. Full model-wise comparisons and experimental details are provided in Appendix~\ref{app:attribution_baseline}.

%% file: sections/limitation.tex
\begin{wrapfigure}{r}{0.3\textwidth}
    \vspace{-1.5em}
    \centering
    \includegraphics[width=1.0\linewidth]{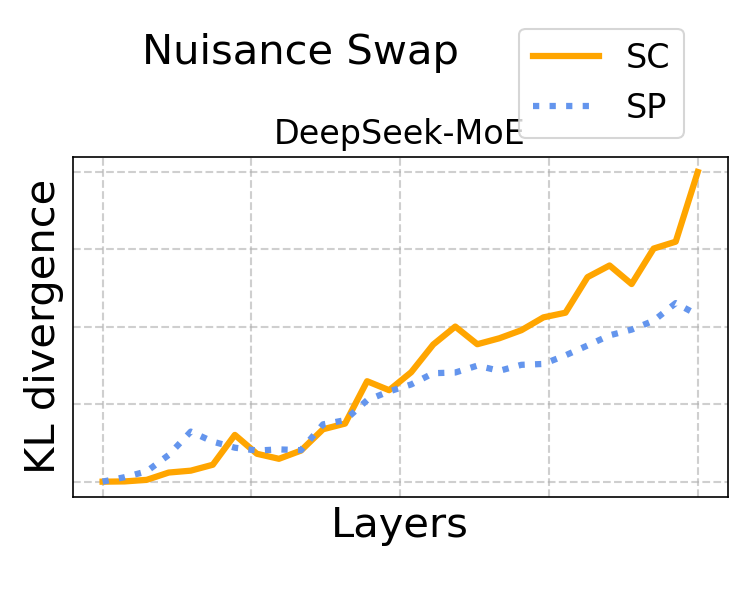}
    \vspace{-2.5em}
    \caption{Per-layer KL divergence under nuisance-component swaps for a model. SP and SC curves increase with depth at comparable rates with no consistent separation. The nuisance subspace likely contains a mixture of surface-form and residual semantic information.}
    \label{fig:nuisance_kl}
    \vspace{-1.5em}
\end{wrapfigure}

Our framework provides a local geometric characterization of semantic invariance rather than a universal decomposition of language representations. Although our evaluation spans 1,400 semantic groups, 57,400 sentences, and nine language models, the decomposition remains dependent on the quality and coverage of the SP and SC perturbations. The dataset is synthetically constructed, and generalization to naturally occurring, multilingual, and long-context inputs remains to be systematically evaluated.

While the invariant subspace $z_{\mathrm{inv}}$ exhibits strong semantic selectivity, the complementary nuisance subspace $z_{\mathrm{nuis}}$ does not achieve complete semantic disentanglement. As shown in Fig.~\ref{fig:nuisance_kl}, swapping nuisance components within the same semantic group produces non-negligible output shifts even for semantic-preserving (SP) donors, indicating that the nuisance representation may retain residual semantic information. Although the two subspaces are orthogonal by construction, the nuisance subspace can still capture a mixture of surface-form variation and residual semantic components. Importantly, this asymmetry does not undermine the primary findings: invariant-component swaps remain stable under SP perturbations while exhibiting strong separation under semantic-changing conditions (Section~\ref{sec:ab_mainpaper}), supporting the conclusion that semantic-sensitive information is concentrated in the invariant subspace. Achieving more complete semantic disentanglement remains an important direction for future work.

%% file: sections/conclusion.tex

We presented a local geometric framework for identifying semantic-invariant features in language model representations through contrastive sensitivity decomposition. Experiments across nine language models demonstrate geometric separation between semantic and nuisance variation, causal sensitivity to semantic changes, and robust zero-shot model attribution. Together, these findings support a local geometric characterization of semantic invariance and its model-specific structure.

%% file: sections/appendix.tex
\hypersetup{
    hidelinks
}

\appendix
\etocdepthtag.toc{appendix}

\section*{Appendix}

\begingroup
\etocsettagdepth{main}{-2}
\etocsettagdepth{appendix}{1}
\etocsetnexttocdepth{1}
\etocsettocstyle{}{}
\tableofcontents
\endgroup

\section{Ablation Study: More Results on Causal Validation via Representation Intervention}
\label{append: between_group}

We provide additional causal intervention results to further analyze the functional role of the discovered invariant and nuisance subspaces across multiple language models. The experiments include both within-group interventions, where donor representations are drawn from the same local semantic neighborhood, and between-group interventions, where donor representations originate from different semantic groups. Together, these analyses evaluate whether the proposed decomposition selectively controls semantic behavior.

\begin{table*}[ht]
\centering
\small
\setlength{\tabcolsep}{4.5pt}
\renewcommand{\arraystretch}{1.2}

\begin{tabular}{c|cc|cc|cc|cc}
\hline

\multirow{2}{*}{Model}
&
\multicolumn{2}{c|}{($A_{\text{nuis}} \leftarrow A_{\text{SP,nuis}}$)}
&
\multicolumn{2}{c|}{($A_{\text{inv}} \leftarrow A_{\text{SP,inv}}$)}
&
\multicolumn{2}{c|}{($A_{\text{nuis}} \leftarrow A_{\text{SC,nuis}}$)}
&
\multicolumn{2}{c}{($A_{\text{inv}} \leftarrow A_{\text{SC,inv}}$)}
\\

&
KL $\downarrow$
&
MSE $\downarrow$
&
KL $\downarrow$
&
MSE $\downarrow$
&
KL $\downarrow$
&
MSE $\downarrow$
&
KL $\uparrow$
&
MSE $\uparrow$
\\

\hline

Mistral-7B-Instruct-v0.3
& 0.124 & 0.581
& 0.036 & 0.401
& 0.109 & 0.612
& 0.341 & 0.951
\\

Meta-Llama-3-8B-Instruct
& 0.138 & 0.630
& 0.040 & 0.434
& 0.114 & 0.646
& 0.360 & 0.990
\\

gemma-7b-it
& 0.131 & 0.602
& 0.042 & 0.455
& 0.121 & 0.670
& 0.382 & 1.044
\\

Qwen2.5-7B-Instruct
& 0.142 & 0.658
& 0.047 & 0.498
& 0.129 & 0.702
& 0.401 & 1.102
\\

glm-4-9b-chat-hf
& 0.118 & 0.541
& 0.034 & 0.386
& 0.111 & 0.594
& 0.337 & 0.918
\\

deepseek-moe-16b-chat
& 0.149 & 0.691
& 0.052 & 0.523
& 0.138 & 0.741
& 0.427 & 1.184
\\

internlm2-chat-7b
& 0.126 & 0.590
& 0.039 & 0.417
& 0.116 & 0.621
& 0.354 & 0.972
\\

Phi-4-mini-instruct
& 0.135 & 0.617
& 0.045 & 0.471
& 0.123 & 0.689
& 0.391 & 1.068
\\

Falcon3-7B-Instruct
& 0.144 & 0.673
& 0.049 & 0.512
& 0.132 & 0.718
& 0.418 & 1.146
\\

\hline

\end{tabular}
\caption{Within-group causal intervention analysis using invariant and nuisance subspace patching across different language models. Interventions are performed using semantic-preserving (SP) and semantic-changing (SC) donors from the same local semantic group. For nuisance swaps ($A_{\mathrm{nuis}} \leftarrow A_{\mathrm{SP,nuis}}$ and $A_{\mathrm{nuis}} \leftarrow A_{\mathrm{SC,nuis}}$), KL divergence and MSE remain relatively low, indicating limited semantic sensitivity. In contrast, invariant swaps with SC donors ($A_{\mathrm{inv}} \leftarrow A_{\mathrm{SC,inv}}$) produce substantially larger deviations than SP donor swaps ($A_{\mathrm{inv}} \leftarrow A_{\mathrm{SP,inv}}$), demonstrating that the discovered invariant subspace selectively controls semantic behavior. Results are averaged across the top-4 invariant layers selected by the largest generalized eigenvalues.}
\label{tab:causal_intervention_analysis}

\end{table*}

\vspace{-0.8em}
\paragraph{Experimental Setup.}
For each semantic group, invariant directions are identified using the discovery subset by solving the generalized eigenvalue problem in Section~\ref{sec: find} and selecting the top-$k$ eigenvectors associated with the largest generalized eigenvalues. The top-4 layers are then selected according to the largest leading eigenvalue $\lambda$, corresponding to layers exhibiting the strongest semantic--nuisance separation. All intervention experiments are performed exclusively on the held-out validation subset, with no overlap between discovery and evaluation perturbations.

At each selected layer, token-level hidden states are decomposed into invariant and nuisance components through orthogonal projection onto the discovered subspaces:
\[
h = h_{\mathrm{inv}} + h_{\mathrm{nuis}}.
\]
Interventions are applied directly to hidden states before propagating the modified activations through the remaining frozen transformer layers.

For within-group interventions, donor representations are sampled from either semantic-preserving (SP) or semantic-changing (SC) perturbations within the same local semantic group. For between-group interventions, donor representations are drawn from a different semantic group, introducing a stronger semantic mismatch between anchor and donor.

We independently replace either the invariant or nuisance component while keeping the complementary component fixed. Invariant interventions are defined as
\[
h^{\mathrm{int}}
=
h^{(a)}
+
\alpha
\left(
h_{\mathrm{inv}}^{(d)}
-
h_{\mathrm{inv}}^{(a)}
\right),
\]
while nuisance interventions are defined as
\[
h^{\mathrm{int}}
=
h^{(a)}
+
\left(
h_{\mathrm{nuis}}^{(d)}
-
h_{\mathrm{nuis}}^{(a)}
\right).
\]

Model behavior before and after intervention is evaluated using KL divergence between next-token distributions and logit-space Mean Squared Error (MSE).

\begin{table}[h]
\centering
\small
\setlength{\tabcolsep}{5pt}
\renewcommand{\arraystretch}{1.2}
\begin{tabular}{c|cc|cc|cc}
\hline
\multirow{2}{*}{Model} 
& \multicolumn{2}{c|}{Group 1}
& \multicolumn{2}{c|}{Group 4}
& \multicolumn{2}{c}{Group 6} \\

& MSE $\downarrow$ & KL $\downarrow$
& MSE $\downarrow$ & KL $\downarrow$
& MSE $\downarrow$ & KL $\downarrow$ \\
\hline

Mistral-7B 
& 0.021 & 0.084 
& 0.024 & 0.091 
& 0.022 & 0.087 \\

Llama-3-8B 
& 0.018 & 0.072 
& 0.020 & 0.079 
& 0.019 & 0.075 \\

Qwen2.5-7B
& 0.019 & 0.078
& 0.003 & 0.033
& 0.020 & 0.080 \\

Gemma-7B
& 0.020 & 0.019
& 0.061 & 0.014
& 0.021 & 0.083 \\

Phi-4-mini
& 0.024 & 0.031
& 0.026 & 0.042
& 0.025 & 0.097 \\

GLM-4-9B
& 0.022 & 0.067
& 0.028 & 0.088
& 0.023 & 0.090 \\

DeepSeek-MoE-16B
& 0.025 & 0.081
& 0.030 & 0.095
& 0.027 & 0.102 \\

InternLM2-7B
& 0.023 & 0.074
& 0.027 & 0.082
& 0.024 & 0.088 \\

Falcon3-7B
& 0.026 & 0.089
& 0.031 & 0.099
& 0.029 & 0.105 \\

\hline
\end{tabular}
\vspace{0.5em}
\caption{Between-group nuisance-component intervention evaluated using Mean Squared Error (MSE) and KL divergence across models and semantic groups. The invariant component is held fixed while the nuisance component is replaced using a donor from a different semantic group. Low MSE and KL indicate that modifying the nuisance subspace induces only limited changes in the output, suggesting that it does not encode primary semantic content. Results are computed using invariant zones defined by top-4 layers selected via generalized eigenvalue analysis.}
\label{tab:between_group_nuis}
\vspace{-2.0em}
\end{table}

\paragraph{Results.}
Tables~\ref{tab:causal_intervention_analysis}, \ref{tab:between_group_nuis}, and~\ref{tab:between_group_inv} summarize the intervention results across all evaluated models.

A consistent asymmetry emerges between invariant and nuisance interventions. For nuisance swaps, both within-group and between-group interventions generally produce relatively small KL divergence and MSE values. Even when nuisance components are replaced using donors from different semantic groups, the resulting output perturbation remains limited. This indicates that the nuisance subspace does not encode primary semantic identity and has comparatively weak causal influence on semantic behavior.

In contrast, invariant interventions produce substantially larger deviations, particularly when donor representations originate from semantically different inputs. Within-group interventions using SP donors produce relatively small changes, while SC donors induce much larger deviations. This separation becomes even more pronounced under between-group interventions, where replacing the invariant component consistently produces strong shifts in model outputs across all architectures.

Importantly, these trends remain highly consistent across model families, layers, and semantic groups. The results therefore provide complementary causal evidence that the discovered invariant subspace encodes semantically meaningful directions governing model behavior, whereas the nuisance subspace primarily captures semantic-preserving variability and non-selective residual variation.

\begin{table}[h]
\centering
\small
\setlength{\tabcolsep}{5pt}
\renewcommand{\arraystretch}{1.2}
\begin{tabular}{c|cc|cc|cc}
\hline
\multirow{2}{*}{Model} 
& \multicolumn{2}{c|}{Group 1}
& \multicolumn{2}{c|}{Group 4}
& \multicolumn{2}{c}{Group 6} \\

& MSE $\uparrow$ & KL $\uparrow$
& MSE $\uparrow$ & KL $\uparrow$
& MSE $\uparrow$ & KL $\uparrow$ \\
\hline

Mistral-7B 
& 0.962 & 0.974 
& 0.968 & 0.978 
& 0.965 & 0.976 \\

Llama-3-8B 
& 0.975 & 0.986 
& 0.978 & 0.988 
& 0.976 & 0.987 \\

Qwen2.5-7B
& 0.971 & 0.984
& 0.982 & 0.991
& 0.972 & 0.985 \\

Gemma-7B
& 0.966 & 0.979
& 0.980 & 0.986
& 0.967 & 0.980 \\

Phi-4-mini
& 0.969 & 0.981
& 0.971 & 0.984
& 0.970 & 0.982 \\

GLM-4-9B
& 0.968 & 0.983
& 0.973 & 0.986
& 0.969 & 0.984 \\

DeepSeek-MoE-16B
& 0.970 & 0.985
& 0.975 & 0.988
& 0.972 & 0.986 \\

InternLM2-7B
& 0.969 & 0.984
& 0.974 & 0.987
& 0.971 & 0.985 \\

Falcon3-7B
& 0.972 & 0.986
& 0.976 & 0.989
& 0.974 & 0.987 \\

\hline
\end{tabular}
\vspace{0.5em}
\caption{Between-group invariant-component intervention evaluated using Mean Squared Error (MSE) and KL divergence across models and semantic groups. The nuisance component is held fixed while the invariant component is replaced using a donor from a different semantic group. High values indicate large and consistent output deviation, suggesting that the invariant subspace encodes semantic identity. Results are computed using invariant zones defined by top-4 layers selected via generalized eigenvalue analysis.}
\label{tab:between_group_inv}
\vspace{-2.0em}
\end{table}

\paragraph{Discussion.}
The contrast between within-group and between-group interventions further clarifies the role of the learned subspaces. Within-group interventions evaluate local semantic stability under paraphrastic variation, while between-group interventions introduce stronger semantic displacement with minimal lexical or structural overlap.

The observed asymmetry between invariant and nuisance interventions suggests that the proposed decomposition captures structured semantic geometry rather than superficial activation statistics. Perturbations in the invariant subspace induce systematic semantic changes, whereas perturbations in the nuisance subspace produce comparatively weak and non-selective effects. Together, these findings support a causal interpretation of invariant zones as semantically meaningful latent directions that persist across models and semantic contexts.

\section{Extended Discussion on Limitation: Diagnostic Analysis of Nuisance Subspace}
\label{app:nuisance_swap}

\paragraph{Token-Level Intervention Protocol.}
We follow the same token-level intervention framework described in Appendix~\ref{append: between_group}. For each input, we extract the hidden state at the final token position and apply the decomposition $h = h_{\mathrm{inv}} + h_{\mathrm{nuis}}$. Interventions are performed at layer $\ell$ by modifying one component while keeping the other fixed, followed by forward propagation to obtain output logits.

\paragraph{Nuisance Swap Intervention.}
To analyze the behavior of the nuisance subspace, we perform interventions that replace the nuisance component of an anchor representation with that of a donor, while keeping the invariant component unchanged. Donor inputs are drawn from either semantic-preserving (SP) or semantic-changing (SC) sets, and the effect is measured using KL divergence between the original and intervened output distributions.

\begin{figure}[h]
    \centering
    \vspace{-1.0em}
    \includegraphics[width=0.8\linewidth]{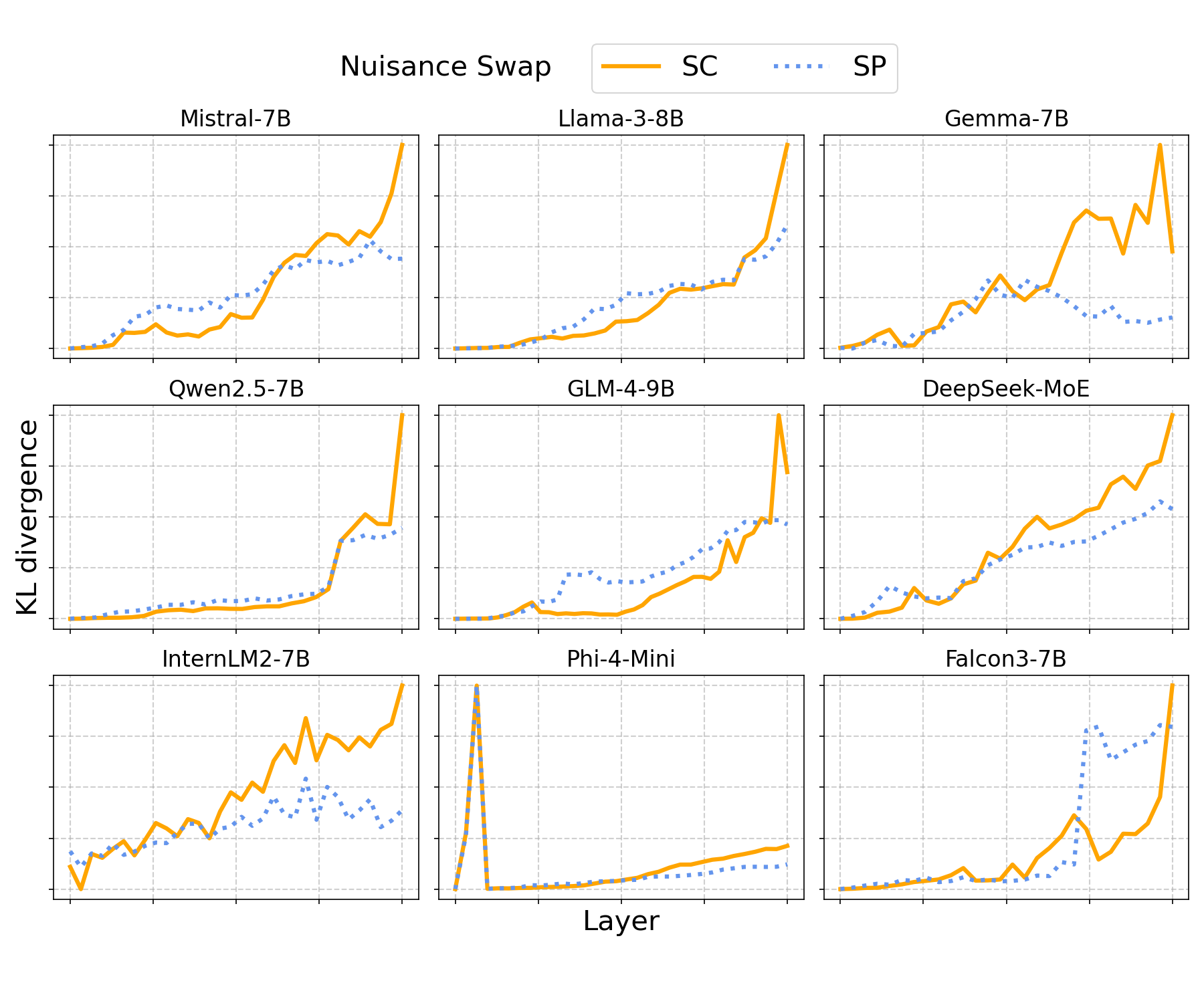}
    \vspace{-2.0em}
    \caption{Per-layer KL divergence for nuisance component swaps across all nine models ($k=32$, $\alpha=1.0$), arranged as a $3\times3$ grid. Orange (solid) curves show SC donor swaps and blue (dotted) curves show SP donor swaps. Both curves increase at comparable rates with no consistent SC-SP separation, indicating that the nuisance subspace does not selectively encode semantic content.}
    \label{fig:nuisance_kl_all}
    \vspace{-1.0em}
\end{figure}

\paragraph{Observations.}
Figure~\ref{fig:nuisance_kl_all} shows the per-layer KL divergence under nuisance swaps across all nine models. In contrast to invariant interventions, SP and SC curves exhibit comparable magnitudes and similar growth patterns across layers, with no consistent separation between the two conditions.

This behavior indicates that perturbations in the nuisance subspace affect model outputs regardless of semantic consistency. Unlike the invariant subspace, where changes align with semantic differences, nuisance perturbations produce output shifts that are not selectively tied to semantic variation.

\paragraph{Diagnostic Interpretation.}
These results provide insight into the limitation observed in Section~5. The decomposition $h = h_{\mathrm{inv}} + h_{\mathrm{nuis}}$ is derived from a data-driven generalized eigenvalue problem that separates directions based on relative sensitivity to semantic-preserving and semantic-changing perturbations. However, geometric orthogonality does not imply complete semantic disentanglement, and the nuisance component may retain residual semantic information.

From a geometric perspective, the tangent-space decomposition 
$
T_z \mathcal{M} = T_z^{\mathrm{semantic}} \oplus T_z^{\mathrm{nuisance}}
$
is only approximately realized in learned representations. Finite-sample estimation, nonlinearity of the representation manifold, and imperfect perturbation design introduce mixing between the two subspaces. As a result, the nuisance component may retain weak projections onto semantic directions. 

This explains why nuisance swaps produce non-negligible output changes even for SP donors: although dominant semantic directions are removed, residual semantic information persists in the nuisance subspace. Consequently, interventions on $h_{\mathrm{nuis}}$ do not isolate purely non-semantic variation.

\vspace{-0.8em}
\paragraph{Implications.}
This asymmetry highlights a key property of the proposed framework. The invariant subspace provides a selective encoding of semantic information, as demonstrated by its strong causal effect under SC interventions and stability under SP perturbations. In contrast, the nuisance subspace is non-selective: it captures a mixture of surface-form variation and residual semantic components, leading to output changes that are insensitive to semantic alignment.

Importantly, this limitation does not invalidate the main conclusions. Rather, it clarifies that the proposed decomposition should be interpreted as a dominant-direction separation rather than a perfectly disentangled factorization. The invariant subspace concentrates semantic structure, while the nuisance subspace captures complementary but partially entangled variation.

\vspace{-0.8em}
\paragraph{Summary.}
Together with the within-group and between-group intervention results, this analysis supports a consistent causal picture: semantic information is concentrated in invariant directions that govern model behavior, while nuisance directions contribute secondary, non-selective variation. Achieving more complete semantic disentanglement of the nuisance decomposition remains an important direction for future work.

\section{Robustness of Invariant Interventions Across Subspace Dimensions}
\label{app:all_groups}

This section evaluates the robustness of the proposed decomposition across models, semantic groups, and subspace dimensions $k \in \{1, 10, 32\}$. We report  within-group per-layer KL divergence under invariant and nuisance component interventions, averaged across all concept groups. The intervention protocol follows Section~4.4, where either the invariant or nuisance component of an anchor representation is replaced using a donor drawn from the SP or SC set.

\begin{figure}[ht]
    \centering
    \vspace{-1.0em}
    \includegraphics[width=0.8\linewidth]{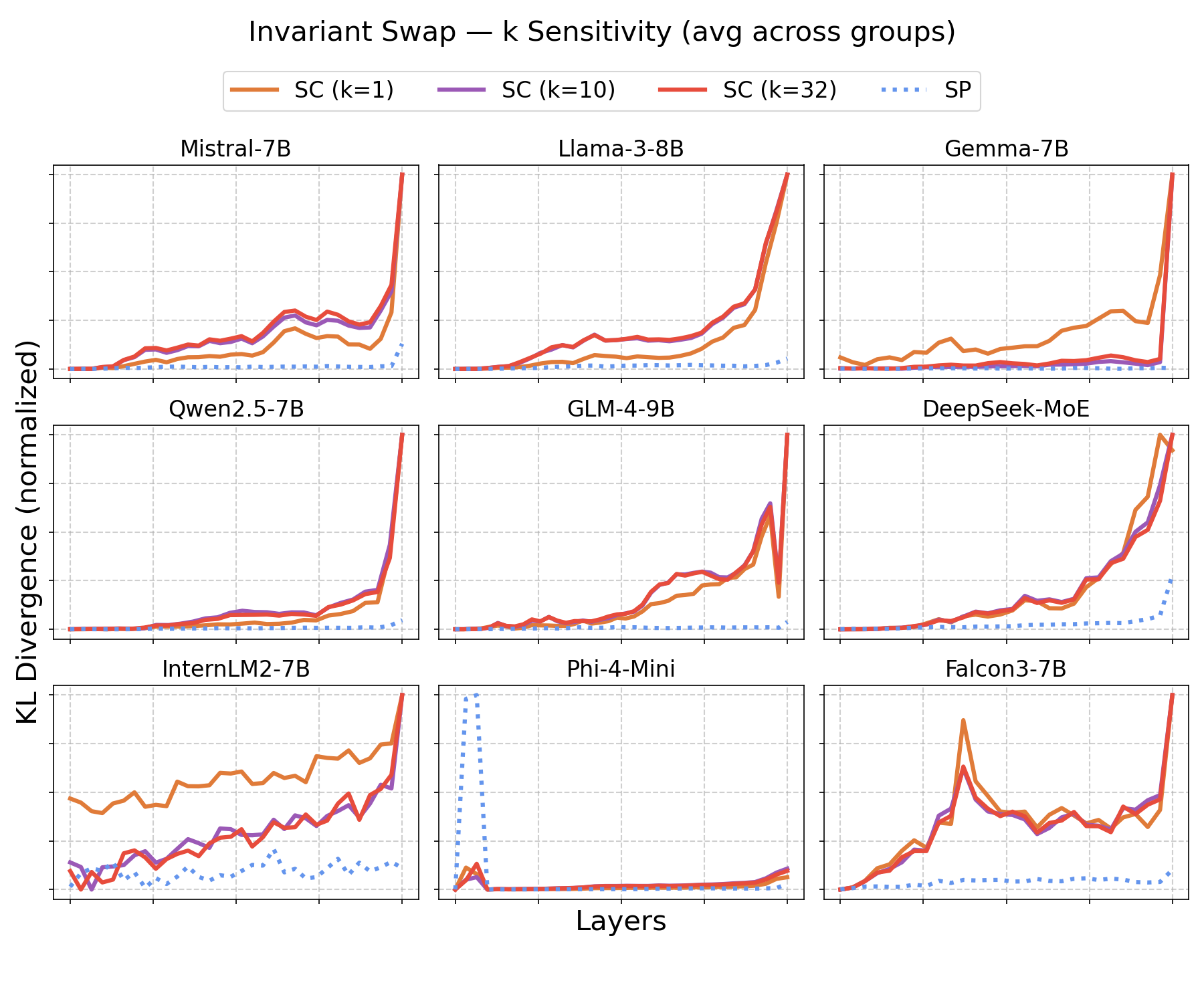}
    \vspace{-2.0em}
    \caption{Per-layer KL divergence for invariant component swaps across all nine models, averaged across all 1400 concept groups ($\alpha=1.0$). Each subplot shows three solid curves corresponding to $k \in \{1, 10, 32\}$ for SC donors, and one dotted curve for SP. SC curves increase consistently with depth while the SP curve remains near zero, suggesting that the invariant subspace selectively encodes semantic content. The three SC curves closely overlap, indicating insensitivity to the choice of $k$.}
    \label{fig:inv_k_comparison}
    \vspace{-1.0em}
\end{figure}

\begin{figure}[ht!]
    \centering
    \vspace{-1.0em}
    \includegraphics[width=0.8\linewidth]{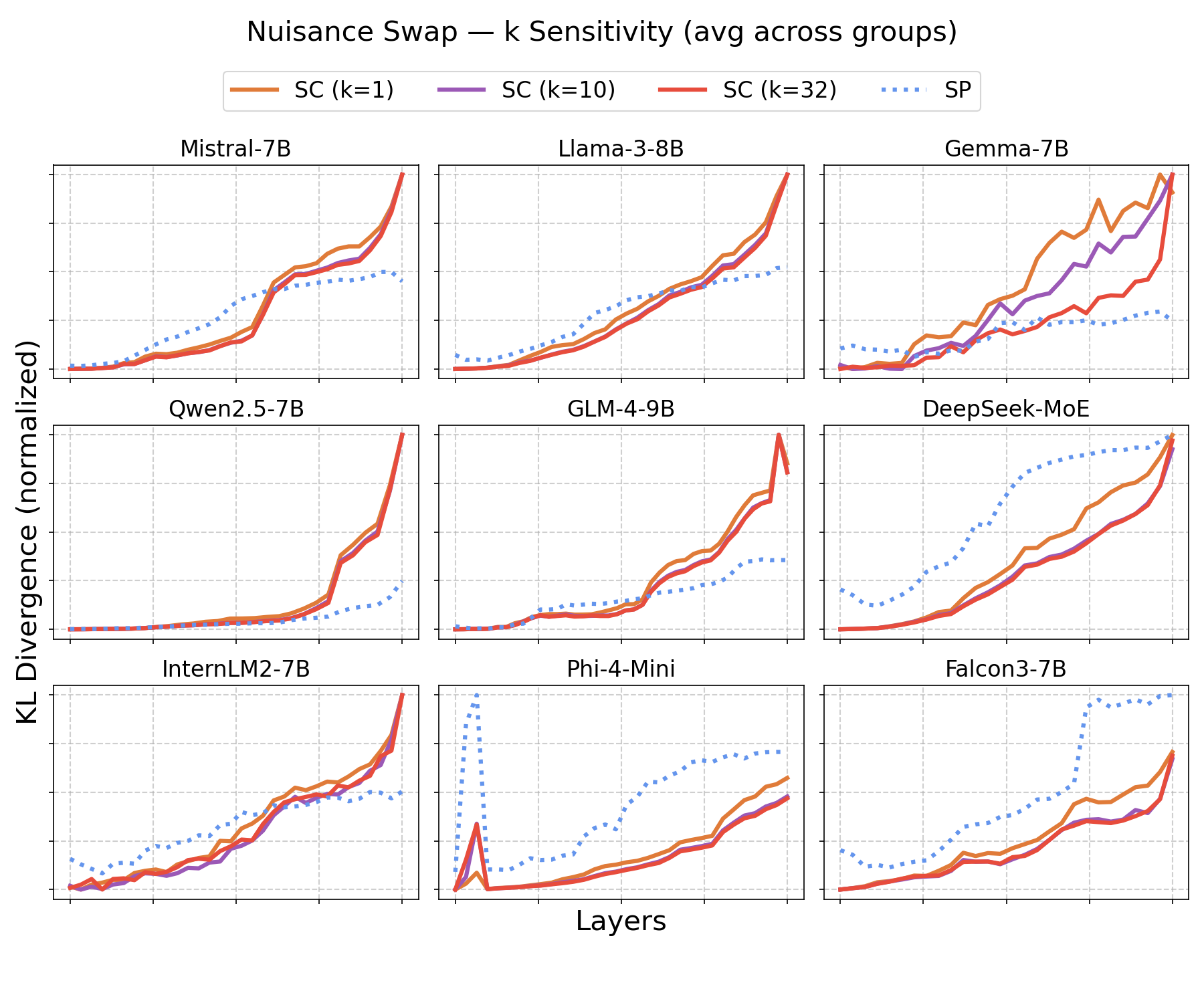}
    \vspace{-2.0em}
    \caption{Per-layer KL divergence for nuisance component swaps across all nine models, averaged across all 1400 concept groups ($\alpha=1.0$). Each subplot shows three solid curves for SC donors at $k \in \{1, 10, 32\}$ and one dotted curve for SP donors. Both SC and SP curves increase at comparable rates with no consistent separation, indicating that the nuisance subspace does not selectively encode semantic content regardless of the choice of $k$.}
    \label{fig:nuis_k_comparison}
    \vspace{-1.0em}
\end{figure}

Figures~\ref{fig:inv_k_comparison} and~\ref{fig:nuis_k_comparison} present results for invariant and nuisance swaps, respectively. Each figure is arranged as a $3 \times 3$ grid with one subplot per model. Solid curves correspond to SC donor swaps for different values of $k$, and the dotted curve corresponds to SP donor swaps.

\paragraph{Interpretation.}
For invariant swaps (Fig.~\ref{fig:inv_k_comparison}), SC curves increase consistently with depth across all models and all values of $k$, while SP curves remain near zero. This confirms that perturbations in the invariant subspace induce semantically meaningful changes, whereas substitutions with semantically equivalent content leave model outputs largely unchanged. The separation between SC and SP conditions becomes most pronounced in mid-to-late layers, aligning with the invariant zones identified in Section~\ref{sec: localize}.

For nuisance swaps (Fig.~\ref{fig:nuis_k_comparison}), both SC and SP curves increase at comparable rates with no consistent separation. This behavior corroborates the diagnostic analysis in Appendix~\ref{app:nuisance_swap}, indicating that the nuisance subspace does not selectively encode semantic content. Instead, it contributes non-selective variation that affects outputs regardless of semantic alignment.

Together, these results reinforce a consistent causal interpretation: invariant components control semantically structured behavior, while nuisance components introduce secondary, non-selective perturbations.

\paragraph{Insensitivity to Subspace Dimension $k$.}
Across both invariant and nuisance interventions, the SC curves for $k \in \{1, 10, 32\}$ closely overlap, indicating that the results are largely insensitive to the choice of subspace dimension. This robustness suggests that the dominant invariant structure is concentrated in a low-dimensional subspace, and that the leading generalized eigenvectors already capture the primary semantic directions.

This observation is consistent with the eigenvalue spectrum in Section~\ref{sec: localize}, where a small number of directions exhibit significantly larger eigenvalues. Increasing $k$ primarily introduces weaker components with diminishing influence on downstream behavior.

From a theoretical perspective, the framework requires only that a subspace exists that separates semantic-preserving and semantic-changing variation, rather than a precise specification of its dimensionality. The empirical stability across $k$ therefore indicates that the identified invariant structure reflects an intrinsic geometric property of the representation, rather than a result of hyperparameter tuning.

\paragraph{Model-Level Variation.}
While the overall pattern is consistent across architectures, Phi-4-Mini exhibits a minor deviation under invariant swaps, with slightly elevated SP responses in early-to-mid layers. This may be attributed to its smaller parameter scale, suggesting that semantic abstraction may be less sharply localized across depth. Despite this variation, the SC–SP separation remains evident, and the overall qualitative behavior is consistent with larger models. We report this as a tentative observation for future investigation.

\section{Comparison with Variance-Based and Contrastive Baselines} 
\label{app:comparison}

\subsection{Attribution Baseline Comparison}
\label{app:attribution_baseline}

\paragraph{Experimental Setup.}

To evaluate whether invariant zone signatures capture model-specific semantic structure beyond trivial architectural separability, we compare the proposed invariant representation (Z$_{\mathrm{inv}}$) against two baselines: a PCA-based latent decomposition and contrastive PCA (cPCA). PCA provides a strong variance-maximizing control that preserves dominant activation energy without explicitly separating semantic-preserving and semantic-changing variation. cPCA, in contrast, explicitly contrasts the two: given the SC covariance $S_{\mathrm{sc}}$ and the SP covariance $S_{\mathrm{sp}}$, it extracts the leading eigenvectors of $S_{\mathrm{sc}} - \alpha S_{\mathrm{sp}}$, where $\alpha$ controls the penalty on variation shared with the semantic-preserving background. 

For each model family, invariant signatures are constructed from the invariant zones identified through the generalized eigenvalue formulation in Section~3.2,
while the PCA and cPCA 
baselines are computed using the same hidden representations and the same subspace dimensionality.
Attribution is performed using held-out validation perturbations only, ensuring that discovery perturbations used for invariant zone identification are never reused during evaluation.

We evaluate three settings: (1) original base models, (2) supervised fine-tuned variants, and (3) knowledge-distilled variants. Fine-tuned and distilled checkpoints are described in Appendix~\ref{app:model_variants}. In all cases, attribution is performed using invariant signatures extracted from the original base model without re-estimating invariant zones for adapted variants. This design tests whether invariant geometric structure remains discriminative under model adaptation.









\begin{table}[!ht]
\centering
\small
\vspace{-1.0em}
\setlength{\tabcolsep}{5pt}
\renewcommand{\arraystretch}{1.2}
\begin{tabular}{l|ccc|ccc|ccc}
\hline
\multirow{2}{*}{Model}
& \multicolumn{3}{c|}{Base Model}
& \multicolumn{3}{c|}{Fine-Tuning}
& \multicolumn{3}{c}{Knowledge Distillation} \\

& PCA & cPCA & Z$_{\text{inv}}$
& PCA & cPCA & Z$_{\text{inv}}$
& PCA & cPCA & Z$_{\text{inv}}$ \\

\hline

Llama-3.1-8B
& 87.4 & 89.7 & \textbf{96.8}
& 85.1 & 87.4 & \textbf{95.3}
& 82.7 & 84.8 & \textbf{93.9} \\

Mistral-7B
& 86.2 & 87.5 & \textbf{95.7}
& 84.6 & 88.2 & \textbf{94.8}
& 81.9 & 84.7 & \textbf{92.6} \\

Qwen2.5-7B
& 85.8 & 92.2 & \textbf{95.1}
& 83.4 & 89.1 & \textbf{93.8}
& 80.6 & 86.3 & \textbf{91.7} \\

Gemma-7B
& 84.9 & 87.9 & \textbf{94.2}
& 82.8 & 85.4 & \textbf{92.9}
& 79.8 & 82.8 & \textbf{90.8} \\

\hline
\end{tabular}
\vspace{0.4em}
\caption{Zero-shot model attribution accuracy (\%) using invariant-zone signatures derived from the proposed contrastive decomposition (Z$_{\mathrm{inv}}$), PCA, and contrastive PCA (cPCA) baselines with matched subspace dimensionality. Evaluation is performed across base, fine-tuned, and knowledge-distilled variants using held-out validation perturbations only. The proposed invariant representation consistently outperforms both PCA and cPCA across all settings, suggesting that the identified invariant zones capture model-specific semantic organization beyond dominant activation variance alone.}
\label{tab:model_attribution_comparison}
\end{table} 

\paragraph{Results.}
Table~\ref{tab:model_attribution_comparison} summarizes attribution accuracy across model families and adaptation settings.
The proposed invariant representation consistently outperforms both PCA and cPCA across all experiments, with improvements remaining stable under both fine-tuning and knowledge distillation. cPCA improves over PCA in every setting but remains below Z$_{\mathrm{inv}}$. 

Importantly, PCA already captures dominant activation variance and therefore implicitly preserves architectural and scale-related information.
The consistent improvement of Z$_{\mathrm{inv}}$ over PCA suggests that attribution performance is not solely explained by superficial architectural separability or raw activation statistics. Instead, the contrastive invariant decomposition identifies directions that are simultaneously semantically stable and discriminative across model families.
The remaining gap to cPCA, which also contrasts SC and SP variation, suggests that the improvement is not only due to using contrastive information but may also reflect how the generalized eigenvalue formulation normalizes SC variation by SP variation. 

These results support the interpretation that invariant zones encode persistent model-specific semantic organization rather than merely reflecting global activation magnitude or dimensionality differences. The fact that attribution remains effective under fine-tuning and distillation further suggests that invariant structure captures stable geometric characteristics that survive substantial model adaptation.

\subsection{Projection Energy Comparison}
\label{app:projection_comparison}

\paragraph{Experimental Setup.}
We evaluate whether the proposed decomposition separates semantic and nuisance variation more effectively than variance-based representations across multiple language model families, including LLaMA, Gemma, Mistral, Qwen, and DeepSeek. Hidden states are extracted from every transformer layer using the final-token representation.

For each semantic group, semantic-preserving (SP) and semantic-changing (SC) perturbations are used to construct covariance statistics for the generalized eigenvalue decomposition described in Section~\ref{sec: find}. The nuisance subspace is defined using the same process as in section~\ref{sec: tangetD}. To evaluate semantic leakage into nuisance directions, we compute semantic centroids for two different semantic groups and measure the normalized projection energy of their semantic displacement vector onto the nuisance subspace. Lower nuisance energy indicates stronger semantic--nuisance separation.

We compare the proposed method against a PCA baseline computed from the same hidden representations and matched subspace dimensionality. PCA covariance matrices are constructed from the union of anchor, SP, and SC samples. All evaluations use held-out validation perturbations only, ensuring that discovery perturbations used for invariant zone identification are never reused during evaluation.

\begin{table}[!htbp]
\centering
\footnotesize
\setlength{\tabcolsep}{5pt}
\renewcommand{\arraystretch}{1.1}

\begin{tabular}{c|cc|cc|cc|cc|cc}
\hline
\multirow{2}{*}{Layer} 
& \multicolumn{2}{c|}{LLaMA} 
& \multicolumn{2}{c|}{Gemma} 
& \multicolumn{2}{c|}{Mistral} 
& \multicolumn{2}{c|}{Qwen} 
& \multicolumn{2}{c}{DeepSeek} \\

& Ours & PCA 
& Ours & PCA
& Ours & PCA
& Ours & PCA
& Ours & PCA \\
\hline

0  & 0.0000 & 0.0000 & 0.0000 & 0.0000 & 0.0000 & 0.0000 & 0.0000 & 0.0000 & 0.0000 & 0.0000 \\
1  & 0.0025 & 0.7320 & 0.0055 & 0.6440 & 0.0028 & 0.5702 & 0.0054 & 0.5998 & 0.0031 & 0.5940 \\
2  & 0.0100 & 0.6208 & 0.0022 & 0.5532 & 0.0069 & 0.5607 & 0.0139 & 0.5746 & 0.0016 & 0.5160 \\
3  & 0.0038 & 0.6892 & 0.0047 & 0.5359 & 0.0024 & 0.6096 & 0.0048 & 0.5158 & 0.0030 & 0.5661 \\
4  & 0.0050 & 0.5584 & 0.0054 & 0.3913 & 0.0035 & 0.4085 & 0.0048 & 0.4519 & 0.0021 & 0.5061 \\
5  & 0.0109 & 0.4182 & 0.0031 & 0.3576 & 0.0037 & 0.4069 & 0.0023 & 0.5534 & 0.0034 & 0.4001 \\
6  & 0.0056 & 0.5097 & 0.0039 & 0.3178 & 0.0039 & 0.5340 & 0.0045 & 0.5970 & 0.0036 & 0.3861 \\
7  & 0.0042 & 0.5357 & 0.0026 & 0.5032 & 0.0017 & 0.5070 & 0.0025 & 0.4837 & 0.0014 & 0.3528 \\
8  & 0.0080 & 0.5216 & 0.0040 & 0.4180 & 0.0050 & 0.4622 & 0.0073 & 0.2989 & 0.0034 & 0.3053 \\
9  & 0.0088 & 0.5324 & 0.0038 & 0.3932 & 0.0065 & 0.4638 & 0.0065 & 0.2803 & 0.0056 & 0.2591 \\
10 & 0.0082 & 0.5330 & 0.0058 & 0.3944 & 0.0075 & 0.4595 & 0.0026 & 0.3821 & 0.0037 & 0.2693 \\
11 & 0.0126 & 0.5927 & 0.0054 & 0.3827 & 0.0052 & 0.4674 & 0.0099 & 0.3565 & 0.0019 & 0.3138 \\
12 & 0.0063 & 0.5857 & 0.0052 & 0.3879 & 0.0038 & 0.4737 & 0.0089 & 0.3121 & 0.0032 & 0.3366 \\
13 & 0.0097 & 0.5625 & 0.0086 & 0.3677 & 0.0050 & 0.4850 & 0.0041 & 0.3216 & 0.0061 & 0.3069 \\
14 & 0.0106 & 0.5745 & 0.0043 & 0.3772 & 0.0047 & 0.5006 & 0.0073 & 0.3188 & 0.0052 & 0.3320 \\
15 & 0.0085 & 0.5698 & 0.0063 & 0.3646 & 0.0028 & 0.5248 & 0.0026 & 0.3224 & 0.0050 & 0.2934 \\
16 & -- & -- & 0.0055 & 0.3707 & 0.0057 & 0.5418 & 0.0063 & 0.3205 & 0.0040 & 0.3201 \\

17 & -- & -- & 0.0050 & 0.3902 & 0.0029 & 0.5669 & 0.0069 & 0.3453 & 0.0020 & 0.3524 \\
18 & -- & -- & 0.0058 & 0.3943 & 0.0012 & 0.5632 & 0.0028 & 0.3630 & 0.0067 & 0.3434 \\
19 & -- & -- & 0.0060 & 0.4044 & 0.0032 & 0.5646 & 0.0051 & 0.4171 & 0.0067 & 0.3419 \\
20 & -- & -- & 0.0046 & 0.3911 & 0.0033 & 0.5314 & 0.0065 & 0.4379 & 0.0052 & 0.3554 \\
21 & -- & -- & 0.0030 & 0.4161 & 0.0025 & 0.5444 & 0.0050 & 0.4555 & 0.0035 & 0.3679 \\
22 & -- & -- & 0.0036 & 0.4396 & 0.0025 & 0.5413 & 0.0066 & 0.4401 & 0.0045 & 0.3634 \\
23 & -- & -- & 0.0051 & 0.4168 & 0.0039 & 0.5410 & 0.0036 & 0.4523 & 0.0025 & 0.3919 \\
24 & -- & -- & 0.0029 & 0.4507 & 0.0045 & 0.5403 & 0.0038 & 0.4400 & 0.0047 & 0.3754 \\
25 & -- & -- & 0.0053 & 0.4403 & 0.0046 & 0.5471 & 0.0034 & 0.4550 & 0.0042 & 0.4409 \\
26 & -- & -- & 0.0020 & 0.4168 & 0.0027 & 0.5526 & 0.0032 & 0.4457 & 0.0025 & 0.4843 \\
27 & -- & -- & 0.0051 & 0.4301 & 0.0037 & 0.5485 & 0.0022 & 0.3956 & 0.0017 & 0.4940 \\
28 & -- & -- & -- & -- & 0.0027 & 0.5334 & -- & -- & 0.0039 & 0.5009 \\
29 & -- & -- & -- & -- & 0.0018 & 0.5279 & -- & -- & 0.0022 & 0.5127 \\
30 & -- & -- & -- & -- & 0.0052 & 0.5190 & -- & -- & 0.0028 & 0.5245 \\
31 & -- & -- & -- & -- & 0.0027 & 0.5206 & -- & -- & 0.0026 & 0.5273 \\

\hline
\end{tabular}
\vspace{0.4em}
\caption{Layer-wise comparison of normalized nuisance projection energy ($E_{\mathrm{nuis}}$) between the proposed invariant decomposition and a PCA baseline across multiple language model families. Lower values indicate better semantic--nuisance separation. Evaluation uses held-out validation perturbations only for representative semantic groups ($\mathrm{group}_g=0$, $\mathrm{group}_h=1$). Missing layers are denoted by (---). A cPCA comparison is given in Table~\ref{tab:cpca_layerwise}. } 
\label{tab:full_layerwise}
\vspace{-1.0em}
\end{table}

\paragraph{Results.}
Table~\ref{tab:full_layerwise} reports full layer-wise nuisance projection energy for representative semantic groups ($\mathrm{group}_g=0$, $\mathrm{group}_h=1$). Across all evaluated architectures and layers, the proposed decomposition consistently produces extremely small nuisance projection energy, typically on the order of $10^{-3}$, while PCA yields substantially larger values ranging from approximately $0.3$ to $0.7$.

These results indicate that the proposed SP/SC-aware decomposition isolates semantic structure substantially more effectively than variance-based PCA decomposition. Importantly, PCA already preserves dominant activation variance and global representation statistics, yet still exhibits large semantic leakage into nuisance directions. The strong and consistent gap therefore suggests that semantic separation is not explained solely by unsupervised variance structure.

The proposed method also remains highly stable across shallow and deep transformer layers, whereas PCA exhibits consistently large nuisance projections throughout the network hierarchy. This behavior further supports the interpretation that semantic structure is more effectively characterized through semantic-preserving and semantic-changing perturbation analysis than through variance maximization alone.

\paragraph{Comparison with cPCA.} 
Table~\ref{tab:cpca_layerwise} provides a representative comparison of nuisance projection energy with cPCA, using the same representations, subspace dimensionality, and evaluation protocol. cPCA substantially improves over PCA, indicating that contrastive information is important. Our method achieves the lowest nuisance energy at all evaluated layers. Nuisance energies on the semantic ambiguity stress test are reported in Table~\ref{tab:ambiguity_results} (Appendix~\ref{app:ambiguity}).

\begin{table}[!htbp]
\centering
\small
\setlength{\tabcolsep}{8pt}
\renewcommand{\arraystretch}{1.15}
\begin{tabular}{cccc}
\toprule
Layer & Ours $\downarrow$ & cPCA $\downarrow$ & PCA $\downarrow$ \\
\midrule
1  & \textbf{0.0025} & 0.0323 & 0.7320 \\
4  & \textbf{0.0050} & 0.0676 & 0.5584 \\
8  & \textbf{0.0085} & 0.0691 & 0.8851 \\
12 & \textbf{0.0053} & 0.0590 & 0.9572 \\
16 & \textbf{0.0047} & 0.0152 & 0.9689 \\
\bottomrule
\end{tabular}
\vspace{0.4em}
\caption{Layerwise nuisance projection energy for LLaMA on the expanded dataset for Ours, cPCA, and PCA. Lower values are better; the best result in each row is shown in bold.}
\label{tab:cpca_layerwise}
\vspace{-1.0em}
\end{table} 









\section{Full Dataset Examples}
\label{app:expanded_data}

\paragraph{Dataset Construction.}
The dataset consists of 1,400 semantic groups spanning 13 domains, comprising 57,400 sentences in total. Each group is centered on an anchor query and contains 20 semantic-preserving (SP) variants, 10 soft semantic-changing (SC) variants, and 10 hard SC variants. SP variants introduce lexical substitutions, syntactic restructuring, or changes in writing style while preserving the propositional content of the anchor. Soft SC variants modify a semantic attribute or slot while largely retaining the original topic and sentence structure. Hard SC variants introduce a topic- or domain-level semantic change, producing a more substantial departure from the anchor's meaning.

This construction provides controlled local neighborhoods containing both semantic-preserving and semantic-changing directions. The distinction between soft and hard SC variants enables analysis of semantic changes at different levels, from localized modifications to broader topic shifts. SC examples are additionally annotated with edit-type and source/target-domain metadata to support perturbation-specific analyses.

\paragraph{Discovery and Held-Out Validation Split.}
For each anchor, the 20 SP variants and 20 SC variants are divided equally into two disjoint subsets: a discovery subset and a held-out validation subset. Each subset contains 10 SP variants, 5 soft SC variants, and 5 hard SC variants. The discovery subset is used exclusively to estimate the local covariance operators, identify invariant subspaces through generalized eigenvalue analysis, and select representation layers. The held-out subset is reserved for downstream evaluations, including projection analysis, causal intervention, and model attribution.

The split is performed within each anchor neighborhood, ensuring that subspace identification and evaluation use disjoint perturbation instances while remaining centered on the same local semantic region. This design evaluates whether the identified invariant directions generalize beyond the perturbation instances used for their estimation, without assuming transfer across unrelated semantic neighborhoods.

\paragraph{Representative Examples.}
Table~\ref{tab:dataset_examples} presents representative examples from ten thematic groups. For each anchor, the first row contains SP and SC variants from the discovery subset, while the second row contains variants from the held-out validation subset. The displayed SC examples illustrate localized semantic changes; the full dataset additionally includes hard SC variants sampled from different topics or domains.

\begin{table}[!htbp]
\centering
\small
\setlength{\tabcolsep}{4pt}
\begin{tabular}{p{1.75cm} p{2.25cm} p{4.25cm} p{4.25cm}}
\toprule
\textbf{Group} & \textbf{Anchor Query} & \textbf{SP Variant} & \textbf{SC Variant} \\
\midrule
\multirow{2}{*}{Cats} &
\multirow{2}{=}{Describe the hunting behavior of domestic cats.} &
Describe the hunting habits of domestic cats. &
Describe the sleeping behavior of domestic cats. \\
& & Detail how house cats pursue and capture prey. &
Describe the nocturnal behavior of domestic cats. \\
\addlinespace
\multirow{2}{*}{France} &
\multirow{2}{=}{What is the official language of France?} &
Which language is officially recognized in France? &
What is the capital city of France? \\
& & Identify the language that serves as the official one in France. &
What is the official flower of France? \\
\addlinespace
\multirow{2}{*}{Python} &
\multirow{2}{=}{Why did Python become the dominant language for data science?} &
What factors led to Python becoming the dominant language in data science? &
Why did Python become popular in academia? \\
& & Explain the reasons behind Python's leading position in data science. &
Why did Python become popular for scientific computing? \\
\addlinespace
\multirow{2}{*}{Photosynthesis} &
\multirow{2}{=}{What are the pros and cons of engineering crops for enhanced photosynthesis?} &
What are the advantages and disadvantages of engineering crops for improved photosynthesis? &
What are the pros and cons of engineering crops for drought resistance? \\
& & Discuss the benefits and costs of engineering crops to improve photosynthetic performance. &
What are the pros and cons of engineering crops for improved carbon sequestration? \\
\addlinespace
\multirow{2}{*}{Democracy} &
\multirow{2}{=}{Describe the role of free speech in a democratic society.} &
What role does free speech play in a democratic society? &
Describe the role of free press in a democratic society. \\
& & Explain the function of free expression in democratic systems. &
Describe the role of voting rights in a democratic society. \\
\addlinespace
\multirow{2}{*}{Einstein} &
\multirow{2}{=}{What is Einstein's most famous theory?} &
Which theory is Albert Einstein best known for? &
What is Einstein's nationality? \\
& & Identify the theory for which Albert Einstein is most widely recognized. &
What is Einstein's most famous thought experiment? \\
\addlinespace
\multirow{2}{*}{Climate Change} &
\multirow{2}{=}{Describe the effect of climate change on global sea levels.} &
What impact does climate change have on global sea levels? &
Describe the effect of climate change on Arctic ecosystems. \\
& & Explain how climate change affects sea levels on a global scale. &
Describe the effect of climate change on global monsoon patterns. \\
\addlinespace
\multirow{2}{*}{Calculus} &
\multirow{2}{=}{How do you find the derivative of a composite function?} &
How is the derivative of a composite function calculated? &
How do you evaluate the limit of a function? \\
& & Explain how to differentiate a function composed of other functions. &
How do you compute the second derivative of a function? \\
\addlinespace
\multirow{2}{*}{Meditation} &
\multirow{2}{=}{Describe the cognitive benefits of regular meditation practice.} &
What cognitive benefits are associated with regular meditation practice? &
Describe the physical benefits of regular meditation practice. \\
& & Explain the mental effects of consistent meditation practice. &
Describe the psychological benefits of regular meditation practice. \\
\addlinespace
\multirow{2}{*}{Bread} &
\multirow{2}{=}{What is the primary leavening agent in sourdough bread?} &
What is the main leavening agent in sourdough bread? &
What is the main ingredient in sourdough bread? \\
& & Identify the ingredient responsible for leavening in sourdough bread. &
What is the typical crust texture of sourdough bread? \\
\bottomrule
\end{tabular}
\vspace{0.4em}
\caption{Representative SP and SC variants from 1400 thematic groups. For each anchor, the first row shows discovery-subset examples and the second row shows held-out validation examples. Each group contains 20 SP variants and 20 SC variants (10 soft and 10 hard), divided equally between the two subsets. The table illustrates soft SC changes; hard SC variants introduce larger topic- or domain-level shifts.}
\label{tab:dataset_examples}
\end{table}

\paragraph{Perturbation Coverage and Evaluation Scope.}
The dataset is designed to characterize local representation geometry under controlled semantic variation rather than to exhaustively sample natural language. Within each anchor neighborhood, the SP and SC perturbations provide contrasting directions for estimating and evaluating semantic invariance. For example, an anchor about domestic cats may be paired with a soft SC variant concerning their sleeping behavior or a hard SC variant such as \textit{``Describe the migration patterns of monarch butterflies.''}

Because semantic preservation is not always binary, we further examine clear SP cases (e.g., harmless typos and grammatical reformulations), ambiguous cases involving scope, punctuation, or syntactic attachment, and clear SC cases involving negation, role reversal, or quantifier changes. The corresponding results are reported in Appendix~\ref{app:ambiguity}.

The consistency of geometric measurements and intervention results across nine language models and diverse semantic groups provides evidence that the observed patterns extend beyond individual anchor queries. Statistical reliability is further assessed using 95\% bootstrap confidence intervals over the 1,400 semantic groups, as reported in Appendix~\ref{app:bootstrap} (Table~\ref{tab:intervention_bootstrap_ci}). Nevertheless, the controlled construction does not fully capture the diversity of naturally occurring language variation, particularly multilingual inputs and long-context dependencies, which remain directions for future evaluation.


\section{Robustness to Semantic Ambiguity}
\label{app:ambiguity}

The boundary between SP and SC perturbations can depend on context: changes in surface form may preserve meaning, alter meaning, or introduce ambiguity. We construct an additional stress test with three categories: clear SP perturbations involving harmless typos or grammatical reformulations; ambiguous perturbations involving scope, punctuation, or attachment; and clear SC perturbations involving negation, role reversal, or quantifier changes. Table~\ref{tab:ambiguity_examples} illustrates these categories. We estimate the decomposition from clear SP and SC variation and evaluate its behavior on the stress-test inputs, including ambiguous cases. Using nuisance projection energy, our method achieves lower values than PCA and cPCA in every reported setting (Table~\ref{tab:ambiguity_results}). These results extend the baseline comparison to inputs where surface form and meaning are closely coupled. The decomposition captures dominant semantic-sensitive directions rather than imposing a rigid binary boundary on every perturbation.

\begin{table}[ht]
\centering
\small
\renewcommand{\arraystretch}{1.15}
\begin{tabular}{p{0.12\linewidth} p{0.27\linewidth}
                p{0.27\linewidth} p{0.23\linewidth}}
\toprule
Category & Original & Perturbation & Semantic Effect \\
\midrule
Clear SP
& Alice is taller than Bob.
& Bob is shorter than Alice.
& Preserves the comparison. \\
\addlinespace
Ambiguity
& I saw the man who was holding a telescope.
& I saw the man with a telescope.
& Introduces ambiguity about who used or held the telescope. \\
\addlinespace
Clear SC
& Not every student submitted the assignment.
& No student submitted the assignment.
& Changes partial non-submission to universal non-submission. \\
\bottomrule
\end{tabular}
\vspace{0.4em}
\caption{Illustrative perturbation pairs for the semantic ambiguity stress test.}
\label{tab:ambiguity_examples}
\vspace{-1.0em}
\end{table}

\begin{table}[t]
\centering
\small
\setlength{\tabcolsep}{6pt}
\renewcommand{\arraystretch}{1.15}
\begin{tabular}{lcccc}
\toprule
Setting & Layer & Ours $\downarrow$ & cPCA $\downarrow$
& PCA $\downarrow$ \\
\midrule
Ambiguity & 3 & \textbf{0.0230} & 0.0718 & 0.1170 \\
Ambiguity & 4 & \textbf{0.0084} & 0.0718 & 0.0921 \\
Clear SP  & 3 & \textbf{0.0122} & 0.0389 & 0.2673 \\
Clear SC  & 4 & \textbf{0.0123} & 0.1284 & 0.2843 \\
\bottomrule
\end{tabular}
\vspace{0.4em}
\caption{Nuisance projection energy on the semantic ambiguity stress test, averaged across all evaluated models. Lower values are better; the best result in each row is shown in bold.}
\label{tab:ambiguity_results}
\vspace{-1.0em}
\end{table}

\section{Statistical Uncertainty on the Expanded Dataset}
\label{app:bootstrap}

We estimate uncertainty in the mean intervention effects across 1,400 semantic groups for each model. At each layer, we perform 1,000 bootstrap trials, each sampling 1,400 groups with replacement and recomputing the mean nuisance- and invariant-component intervention metrics. We obtain 95\% confidence intervals from the 2.5th and 97.5th percentiles of the bootstrap distribution. Table~\ref{tab:intervention_bootstrap_ci} reports the across-group means and confidence intervals on the expanded dataset at representative layers (3, 5, and 8). For every model and every reported layer, invariant-component interventions produce larger MSE and KL changes than nuisance-component interventions, with confidence intervals narrow relative to the separation between the two intervention types.

\begin{table}[!htbp]
\centering
\small
\setlength{\tabcolsep}{3pt}
\renewcommand{\arraystretch}{1.15}
\begin{tabular}{p{1.5cm} p{0.75cm} p{2.00cm} p{2.00cm} p{2.00cm} p{2.00cm}}
\toprule
Model & Layer
& Nuisance KL & Invariant KL
& Nuisance MSE & Invariant MSE \\
\midrule
Mistral-7B
& 3
& $0.000294 \pm 0.000022$
& $0.183353 \pm 0.000924$
& $0.001498 \pm 0.000840$
& $0.554108 \pm 0.000398$ \\
& 5
& $0.001740 \pm 0.000084$
& $0.048431 \pm 0.000880$
& $0.001239 \pm 0.000699$
& $0.124954 \pm 0.000296$ \\
& 8
& $0.000535 \pm 0.000022$
& $0.115219 \pm 0.000627$
& $0.002914 \pm 0.000477$
& $0.538303 \pm 0.000858$ \\
\midrule
Llama-3-8B
& 3
& $0.001294 \pm 0.000098$
& $0.010279 \pm 0.000647$
& $0.003071 \pm 0.000194$
& $0.024246 \pm 0.001549$ \\
& 5
& $0.006545 \pm 0.000497$
& $0.039726 \pm 0.001782$
& $0.016768 \pm 0.001773$
& $0.091234 \pm 0.003889$ \\
& 8
& $0.009954 \pm 0.000670$
& $0.109119 \pm 0.003465$
& $0.027611 \pm 0.002654$
& $0.211765 \pm 0.007343$ \\
\midrule
Qwen2.5-7B
& 3
& $0.001692 \pm 0.000049$
& $0.141399 \pm 0.000169$
& $0.002138 \pm 0.000237$
& $0.280972 \pm 0.000876$ \\
& 5
& $0.000498 \pm 0.000021$
& $0.103953 \pm 0.000996$
& $0.001561 \pm 0.000520$
& $0.449803 \pm 0.000882$ \\
& 8
& $0.002991 \pm 0.000074$
& $0.134659 \pm 0.001000$
& $0.001339 \pm 0.000355$
& $0.024437 \pm 0.000509$ \\
\midrule
Gemma-7B
& 3
& $0.000982 \pm 0.000041$
& $0.083551 \pm 0.000415$
& $0.001444 \pm 0.000264$
& $0.528555 \pm 0.000336$ \\
& 5
& $0.000252 \pm 0.000035$
& $0.174981 \pm 0.000732$
& $0.002773 \pm 0.000624$
& $0.362715 \pm 0.000635$ \\
& 8
& $0.001787 \pm 0.000030$
& $0.175533 \pm 0.000438$
& $0.002566 \pm 0.000295$
& $0.143605 \pm 0.000834$ \\
\midrule
Phi-4-mini
& 3
& $0.001135 \pm 0.000073$
& $0.150466 \pm 0.000103$
& $0.002899 \pm 0.000279$
& $0.216349 \pm 0.000892$ \\
& 5
& $0.001609 \pm 0.000055$
& $0.157141 \pm 0.000147$
& $0.002398 \pm 0.000918$
& $0.200817 \pm 0.000420$ \\
& 8
& $0.000629 \pm 0.000069$
& $0.137118 \pm 0.000304$
& $0.002453 \pm 0.000747$
& $0.483504 \pm 0.000699$ \\
\midrule
GLM-4-9B
& 3
& $0.001060 \pm 0.000083$
& $0.025708 \pm 0.000593$
& $0.002141 \pm 0.000409$
& $0.017117 \pm 0.000323$ \\
& 5
& $0.002712 \pm 0.000082$
& $0.077275 \pm 0.000622$
& $0.002567 \pm 0.000122$
& $0.074551 \pm 0.000403$ \\
& 8
& $0.001142 \pm 0.000044$
& $0.181325 \pm 0.000979$
& $0.001591 \pm 0.000384$
& $0.142025 \pm 0.000125$ \\
\midrule
DeepSeek-MoE-16B
& 3
& $0.002609 \pm 0.000038$
& $0.147375 \pm 0.000650$
& $0.002082 \pm 0.000829$
& $0.265895 \pm 0.000227$ \\
& 5
& $0.002764 \pm 0.000089$
& $0.190571 \pm 0.000611$
& $0.002600 \pm 0.000257$
& $0.029917 \pm 0.000208$ \\
& 8
& $0.001402 \pm 0.000064$
& $0.027540 \pm 0.000402$
& $0.001993 \pm 0.000376$
& $0.117297 \pm 0.000493$ \\
\midrule
InternLM2-7B
& 3
& $0.002665 \pm 0.000025$
& $0.163942 \pm 0.000931$
& $0.001603 \pm 0.000531$
& $0.231792 \pm 0.000880$ \\
& 5
& $0.000171 \pm 0.000056$
& $0.106152 \pm 0.000652$
& $0.002800 \pm 0.000651$
& $0.209048 \pm 0.000898$ \\
& 8
& $0.002509 \pm 0.000086$
& $0.037368 \pm 0.000844$
& $0.001217 \pm 0.000796$
& $0.292352 \pm 0.000495$ \\
\midrule
Falcon3-7B
& 3
& $0.000382 \pm 0.000039$
& $0.087830 \pm 0.000734$
& $0.002887 \pm 0.000966$
& $0.117867 \pm 0.000723$ \\
& 5
& $0.001994 \pm 0.000037$
& $0.102695 \pm 0.000221$
& $0.002094 \pm 0.000239$
& $0.392787 \pm 0.000200$ \\
& 8
& $0.001578 \pm 0.000037$
& $0.027555 \pm 0.000999$
& $0.002944 \pm 0.000565$
& $0.310904 \pm 0.000624$ \\
\bottomrule
\end{tabular}
\vspace{0.4em}
\caption{Mean intervention metrics on the expanded dataset across 1,400 semantic groups at layers 3, 5, and 8, with 95\% bootstrap confidence intervals. Results are reported as mean $\pm$ confidence interval half-width.}
\label{tab:intervention_bootstrap_ci}
\vspace{-1.0em}
\end{table}

\section{Effect of Pooling Strategy}
\label{app:pooling}

We compare mean pooling and last-token pooling across four models at layer 3 using the same decomposition and intervention protocol. As shown in Table~\ref{tab:pooling_comparison}, both strategies produce larger SC than SP intervention effects in KL divergence and MSE, preserving semantic selectivity. Last-token pooling consistently yields lower SP disturbance and nuisance projection energy, although its absolute SC effects are also smaller. We therefore retain last-token pooling as a conservative choice that limits sensitivity to meaning-preserving changes while remaining responsive to semantic changes.

\begin{table}[!htbp]
\centering
\small
\setlength{\tabcolsep}{4pt}
\renewcommand{\arraystretch}{1.15}
\begin{tabular}{llccccc}
\toprule
Model & Pooling
& SP KL $\downarrow$
& SC KL $\uparrow$
& SP MSE $\downarrow$
& SC MSE $\uparrow$
& Nuisance Energy $\downarrow$ \\
\midrule
Llama-3-8B & Mean
& 0.001190 & 0.065358 & 0.003095 & 0.123798 & 0.061882 \\
& Last token
& 0.000037 & 0.000876 & 0.000087 & 0.001567 & 0.013685 \\
\midrule
Mistral-7B & Mean
& 0.002501 & 0.027186 & 0.000094 & 0.004563 & 0.061063 \\
& Last token
& 0.000027 & 0.001793 & 0.000089 & 0.000729 & 0.009884 \\
\midrule
Qwen2.5-7B & Mean
& 0.000692 & 0.038182 & 0.000187 & 0.007310 & 0.012205 \\
& Last token
& 0.000087 & 0.000445 & 0.000094 & 0.000774 & 0.006332 \\
\midrule
Gemma-7B & Mean
& 0.001159 & 0.008129 & 0.003581 & 0.007889 & 0.066067 \\
& Last token
& 0.000108 & 0.000355 & 0.000199 & 0.000385 & 0.044512 \\
\bottomrule
\end{tabular}
\vspace{0.4em}
\caption{Comparison of mean pooling and last-token pooling at layer 3 under the same decomposition and intervention protocol.}
\label{tab:pooling_comparison}
\vspace{-1.0em}
\end{table}

\section{Multi-hop Reasoning Validation}
\label{app:multihop}

Multi-hop reasoning requires combining multiple facts or intermediate inferences to reach a conclusion, rather than relying on a single explicit fact. We additionally evaluate 200 multi-hop reasoning examples to test whether the invariant decomposition remains selective when semantic differences require reasoning beyond direct lexical or attribute substitutions. Table~\ref{tab:multihop_perturbation_examples} shows example instances from this dataset. Using the same decomposition and intervention protocol, we compare the effects of replacing semantic and nuisance components. As shown in Table~\ref{tab:multihop_validation}, semantic interventions produce larger MSE and KL changes than nuisance interventions at all reported layers. These results support the conclusion that the discovered invariant directions remain semantically selective in reasoning-intensive settings.

\begin{table}[!htbp]
\centering
\small
\begin{tabular}{p{0.13\linewidth} p{0.65\linewidth} p{0.13\linewidth}}
\toprule
Type & Supporting Facts & Answer \\
\midrule
\multicolumn{3}{p{0.95\linewidth}}{
\textbf{Example 1:} In which country does Maya live?} \\
Anchor &
Maya lives in Brookfield. Brookfield is in Canada. &
Canada \\
SP &
Maya resides in Brookfield. Brookfield is located in Canada. &
Canada \\
SC &
Maya lives in Brookfield. Brookfield is in France. &
France \\
\midrule
\multicolumn{3}{p{0.95\linewidth}}{
\textbf{Example 2:} Which piece of furniture contains the key?} \\
Anchor &
The key is inside a box. The box is inside a cupboard. &
Cupboard \\
SP &
A box holds the key. The box is stored inside a cupboard. &
Cupboard \\
SC &
The key is inside a box. The box is inside a desk. &
Desk \\
\midrule
\multicolumn{3}{p{0.95\linewidth}}{
\textbf{Example 3:} Did Aisha finish before Carlos?} \\
Anchor &
Aisha finished before Ben. Ben finished before Carlos. &
Yes \\
SP &
Ben finished after Aisha but before Carlos. &
Yes \\
SC &
Aisha finished after Ben. Carlos finished before Ben. &
No \\
\bottomrule
\end{tabular}
\vspace{0.4em}
\caption{Illustrative multi-hop examples with nuisance (SP) and semantic (SC) changes. SP variants preserve the reasoning and answer, whereas SC variants change a fact or relationship and alter the answer. The question remains fixed within each example.}
\label{tab:multihop_perturbation_examples}
\vspace{-1.0em}
\end{table}

\begin{table}[!htbp]
    \centering
    \small
    \begin{tabular}{ccccc}
        \toprule
        Layer
        & Nuisance MSE
        & Semantic MSE
        & Nuisance KL
        & Semantic KL \\
        \midrule
        3 & 0.143 & 0.970 & 0.064 & 0.353 \\
        4 & 0.130 & 0.908 & 0.069 & 0.360 \\
        5 & 0.107 & 0.832 & 0.080 & 0.391 \\
        \bottomrule
    \end{tabular}
\vspace{0.4em}
\caption{Intervention results on the additional multi-hop reasoning dataset, averaged across all evaluated models. Semantic-component interventions produce larger changes than nuisance-component interventions at each reported layer.}
\label{tab:multihop_validation}
\vspace{-1.0em}
\end{table}

\section{Implementation Details}
\label{appendix: implementation}


\subsection{Generalized Eigenvalue Regularization and Whitening Equivalence}
\label{app:whitening}

Since the number of perturbations is much smaller than the representation dimension, the empirical semantic-preserving covariance $S_{\mathrm{SP}}$ is typically rank-deficient. We therefore introduce ridge regularization $B=S_{\mathrm{SP}}+\epsilon I$, with $\epsilon>0$ ($\epsilon$ is set to a small fraction, e.g., $10^{-4}$), and solve
\begin{equation}
S_{\mathrm{SC}}v=\lambda Bv.
\end{equation}
Because $B$ is symmetric positive definite, its inverse square root $B^{-1/2}$ exists. Substituting $u=B^{1/2}v$ yields the equivalent standard eigenvalue problem
\begin{equation}
B^{-1/2}S_{\mathrm{SC}}B^{-1/2}u=\lambda u,
\qquad v=B^{-1/2}u.
\end{equation}
Thus, our generalized eigenvalue formulation is equivalent to whitening the representation space with respect to regularized SP variation before identifying the dominant SC-sensitive directions. The regularization ensures numerical stability when $S_{\mathrm{SP}}$ is ill-conditioned. The resulting generalized eigenvectors are subsequently Euclidean-orthonormalized, as described in Section~\ref{sec: find}, to construct the invariant projection operator.

\subsection{Cross-Layer Relationship of Invariant Subspaces}

Invariant subspaces are estimated independently at each transformer layer, without assuming a fixed basis across depth. Since adjacent layers are connected through nonlinear transformations, their local invariant directions may be approximately related through the layer Jacobian, i.e., $U_{\ell+1}^{\mathrm{inv}}\approx\operatorname{orth}(J_\ell U_\ell^{\mathrm{inv}})$. However, this relationship is input-dependent and is not imposed or evaluated in the present framework. Characterizing cross-layer subspace transport remains an interesting direction for future work.

\subsection{Representation Extraction}
For each sentence, we extract the hidden state at the final token position from every transformer layer using forward hooks. In causal language models, the final token attends to all preceding tokens and aggregates contextual information from the full input, making it the most semantically rich position for next-token prediction. This yields one $d$-dimensional representation per sentence per layer, where $d$ is the model's hidden dimension (4096 for most models evaluated). All representations are extracted in \texttt{float32} precision.

\subsection{Implementation Details for Nuisance Energy Evaluation.}
For nuisance energy evaluation, we compute semantic displacement between all ordered semantic group pairs $(g,h)$ with $g \neq h$. Given $N$ semantic groups, this produces $N(N-1)$ ordered semantic displacement pairs. In our experiments, we use a predefined set of semantic groups to generate semantic displacement pairs for each model and layer.

Nuisance energy is evaluated independently at every transformer layer, including the embedding layer and all subsequent transformer blocks. Unless otherwise specified, the invariant subspace dimensionality is fixed to $k=32$, and the nuisance subspace is defined as its Euclidean orthogonal complement.
Hidden representations are extracted from the final-token embedding at each layer and converted to floating-point representations prior to covariance estimation. 
The sensitivity matrices are computed from anchor-relative displacement vectors $\Delta z=z-z_{\mathrm{orig}}$ as defined in Section~\ref{sec: find}. 
Semantic displacement vectors are not additionally normalized before projection. Instead, normalization is incorporated implicitly through the energy-ratio formulation, where projected nuisance energy is divided by the total displacement energy.

Table~\ref{tab:energy} reports some representative groups. Across all evaluated models, the proposed method consistently yields very small nuisance projection energy with low variance across semantic groups and layers, indicating stable separation between semantic and nuisance components throughout the transformer hierarchy.

In contrast, variance-based decomposition methods produce substantially larger nuisance projection energy and greater variability, suggesting weaker disentanglement between semantic and nuisance directions. The consistently low nuisance energy observed in our method therefore reflects stable concentration of semantic variation within the invariant subspace rather than numerical artifacts or isolated layer behavior.

\subsection{Compute Resources}
\label{app:compute_resources}

All experiments were conducted on a single NVIDIA RTX 6000 Ada Generation GPU (48 GB VRAM) from a workstation equipped with two such GPUs, an AMD Ryzen Threadripper PRO 7985WX (64 cores), and 252 GB system RAM. A single GPU was used throughout to ensure compatibility with single-GPU reproduction environments. Representation extraction for one model across all layers for one concept group takes approximately 1 minute. The causal intervention experiment takes approximately 3 minutes per model per group. Fine-tuning each base model on the Alpaca dataset for one epoch using LoRA takes approximately 20--40 minutes per model. The total compute for all reported experiments, including eigenvalue analysis, intervention experiments, attribution evaluation, and t-SNE visualization across all models, concept groups, and values of $k$, is estimated at approximately 150 GPU-hours, inclusive of preliminary and repeated runs during development.

\subsection{Fine-Tuned and Distilled Model Variants}
\label{app:model_variants}

To evaluate the robustness of invariant representations under model adaptation, we construct fine-tuned and distilled variants of all nine base models, yielding 18 additional checkpoints (9 fine-tuned, 9 distilled) for a total of 27 model variants evaluated in the attribution experiments.

\begin{table}[!htbp]
\centering
\small
\begin{tabular}{llcc}
\toprule
\textbf{Teacher Model} & \textbf{Student Model} & \textbf{Teacher Params} & \textbf{Student Params} \\
\midrule
Mistral-7B-Instruct-v0.3  & Gemma-2B          & 7B  & 2B   \\
Gemma-7B-IT               & Gemma-2B          & 7B  & 2B   \\
Llama-3-8B-Instruct       & TinyLlama-1.1B-Chat & 8B  & 1.1B \\
Qwen2.5-7B-Instruct       & Qwen-1.5B         & 7B  & 1.5B \\
GLM-4-9B-Chat             & Phi-3-Mini        & 9B  & 3.8B \\
DeepSeek-MoE-16B-Chat     & Phi-3-Mini        & 16B & 3.8B \\
InternLM2-7B-Chat         & Phi-3-Mini        & 7B  & 3.8B \\
Phi-4-Mini-Instruct       & Phi-3-Mini        & 4B  & 3.8B \\
Falcon3-7B-Instruct       & Phi-3-Mini        & 7B  & 3.8B \\
\bottomrule
\end{tabular}
\vspace{0.4em}
\caption{Distilled model variants used for attribution robustness evaluation. Each student is trained on the Alpaca dataset to approximate the output distribution of the corresponding teacher via KL divergence minimization. Fine-tuned variants (not shown) are produced for all nine base models using the same Alpaca dataset with standard supervised fine-tuning.}
\label{tab:model_variants}
\vspace{-1.0em}
\end{table}

\paragraph{Checkpoint Sources.}
All adapted model checkpoints are trained by the authors from the original base model weights obtained from HuggingFace Hub. 
Fine-tuned variants are produced by applying supervised fine-tuning with Low-Rank Adaptation (LoRA) on the Alpaca instruction-following dataset~\cite{alpaca}, which contains approximately 52,000 instruction-response pairs. Fine-tuning uses cross-entropy loss on the response tokens, with the pretrained model weights frozen and only the LoRA adapter parameters updated. 
Distilled variants are produced by training a smaller student model to approximate the output distribution of the corresponding teacher on the same Alpaca dataset, using response-level knowledge distillation where the student minimizes the KL divergence between its logits and those of the teacher. Student architectures are selected based on proximity to the teacher family where possible: Gemma-2B for Mistral and Gemma, TinyLlama-1.1B-Chat for Llama, Qwen-1.5B for Qwen, and Phi-3-Mini for the remaining five models. Table~\ref{tab:model_variants} summarizes the distillation pairings and parameter scales.

\paragraph{Signature Construction and Layer Selection.}
Invariant zone signatures are computed exclusively from the original base model for each model family. The same signatures and layer selections are reused without modification when evaluating fine-tuned and distilled variants. This protocol is intentional: it tests whether the geometric structure discovered in the base model persists under adaptation, rather than re-discovering potentially adapted structure in each variant separately. We acknowledge that this means layer selection is performed on base model representations and may not reflect the optimal layers for adapted variants. However, since attribution is performed by matching a test variant against a base model prototype, the relevant question is whether the base model's invariant zones remain discriminative after adaptation — which is precisely what the results in Table~\ref{tab:model_wise_attribution} evaluate. Layer selection does not use model identity as a signal and is performed independently per model family prior to any attribution comparison, so no identity leakage occurs through this step.

All 27 model variants are evaluated using invariant zone signatures derived solely from the original base model checkpoints. This design isolates whether invariant representations encode persistent geometric structure that survives model adaptation, rather than being specific to a particular training state.


\section{Visualizing the Geometric Separability of Invariant Representations}
\label{app:extra_eval}
We visualize that the invariant subspace identified by our framework yields geometrically separable representations across models. We use t-SNE~\cite{vandermaaten2008} to visualize representation geometry in two settings: our structured custom dataset and an external corpus of queries sampled from MS MARCO~\cite{msmarco}, a large-scale information retrieval benchmark consisting of real web search queries. MS MARCO queries are drawn from a fundamentally different distribution than our custom SP/SC dataset—they are natural, unstructured, and span a broad range of topics without controlled semantic variation—making them a suitable testbed for evaluating whether the invariant subspace generalizes beyond the data used to derive it.

\begin{figure}[h]
    \centering
    \vspace{-0.5em}
    \includegraphics[width=0.7\linewidth]{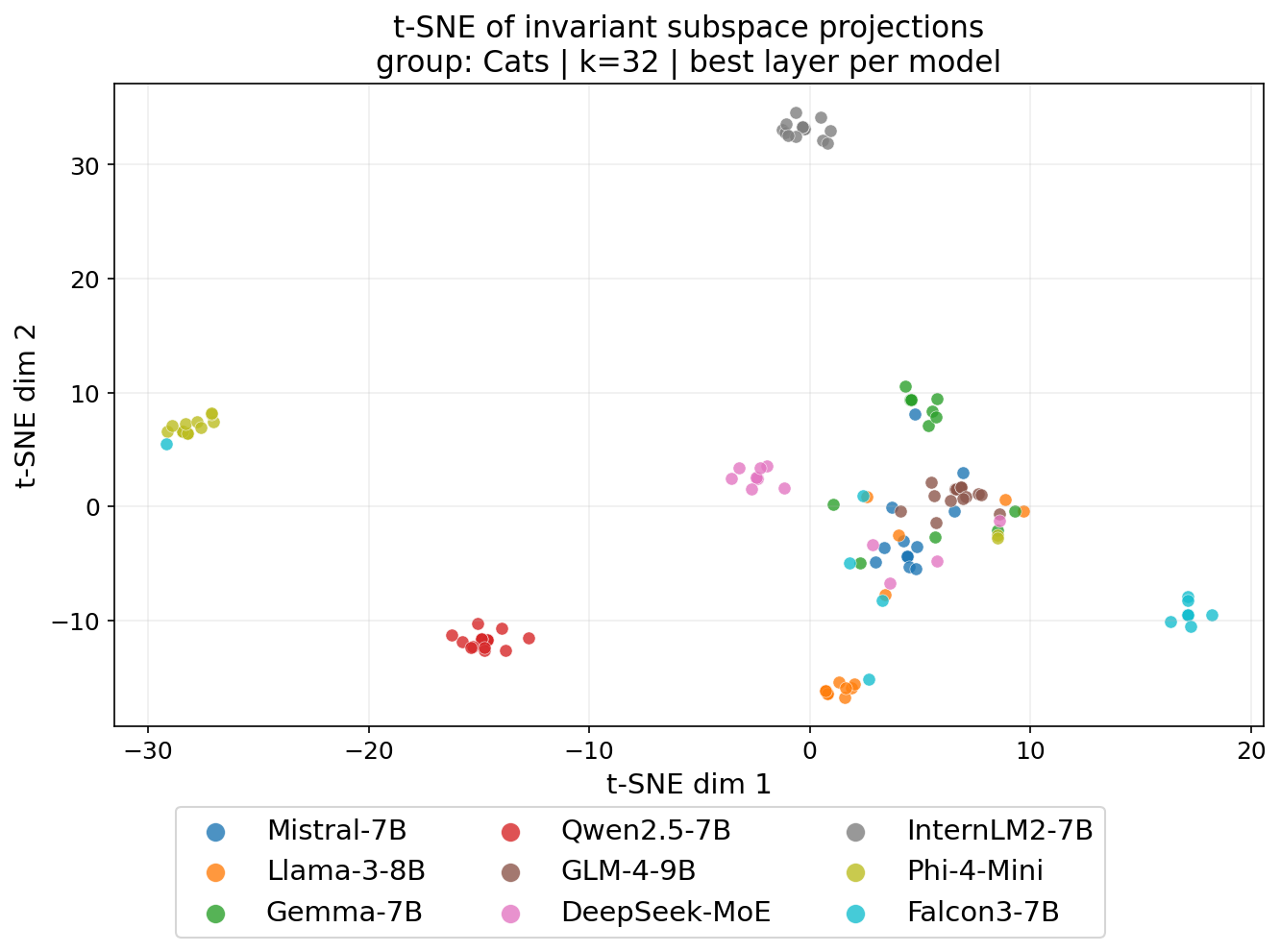}
    \vspace{-0.7em}
    \caption{t-SNE visualization of invariant subspace projections ($k=32$) for all nine models on a single concept group. Each point represents one sentence, colored by model identity.}
    \label{fig:tsne_custom}
    \vspace{-0.5em}
\end{figure}

\paragraph{Custom Dataset.}
Figure~\ref{fig:tsne_custom} shows a t-SNE visualization of invariant subspace projections for all nine models on a single concept group ($k=32$). For each model, the best layer is selected independently as the layer where the top-$k$ generalized eigenvectors account for the largest fraction of invariant signal. To ensure comparability across layers, eigenvalues at each layer are normalized by the sum of positive eigenvalues at that layer. This normalization emphasizes the relative concentration of invariant signal within each layer rather than absolute magnitude, which increases with depth and would otherwise bias selection toward later layers. Hidden states at the selected layer are collected for all sentences in the group (1 anchor, 10 SP, 10 SC), and each representation is projected into the $k$-dimensional invariant coordinate space via $c = U_{\mathrm{inv}}^\top h$. The resulting coordinates are aggregated across models and reduced to 2D using t-SNE for visualization. Despite the limited number of samples per model, the representations form visually separable clusters by model identity, suggesting that the invariant subspace captures model-specific geometric structure.

\paragraph{MS MARCO Queries.}
To evaluate generalization beyond the structured SP/SC dataset, we sample 200 queries from the MS MARCO training split and extract hidden state representations for each model at its best layer. The best layer is selected as the one with the highest mean normalized eigenvalue fraction averaged across all concept groups. This cross-group averaging ensures that the selected layer reflects consistently strong invariant structure across diverse semantic domains rather than being tuned to a single concept. Critically, $U_{\mathrm{inv}}$ is derived from the custom SP/SC dataset and held fixed; it is applied to MS MARCO representations without recomputation, testing whether invariant subspaces discovered on controlled data transfer to natural language queries from a different distribution.

\begin{figure}[h]
    \centering
    \vspace{-0.5em}
    \includegraphics[width=\linewidth]{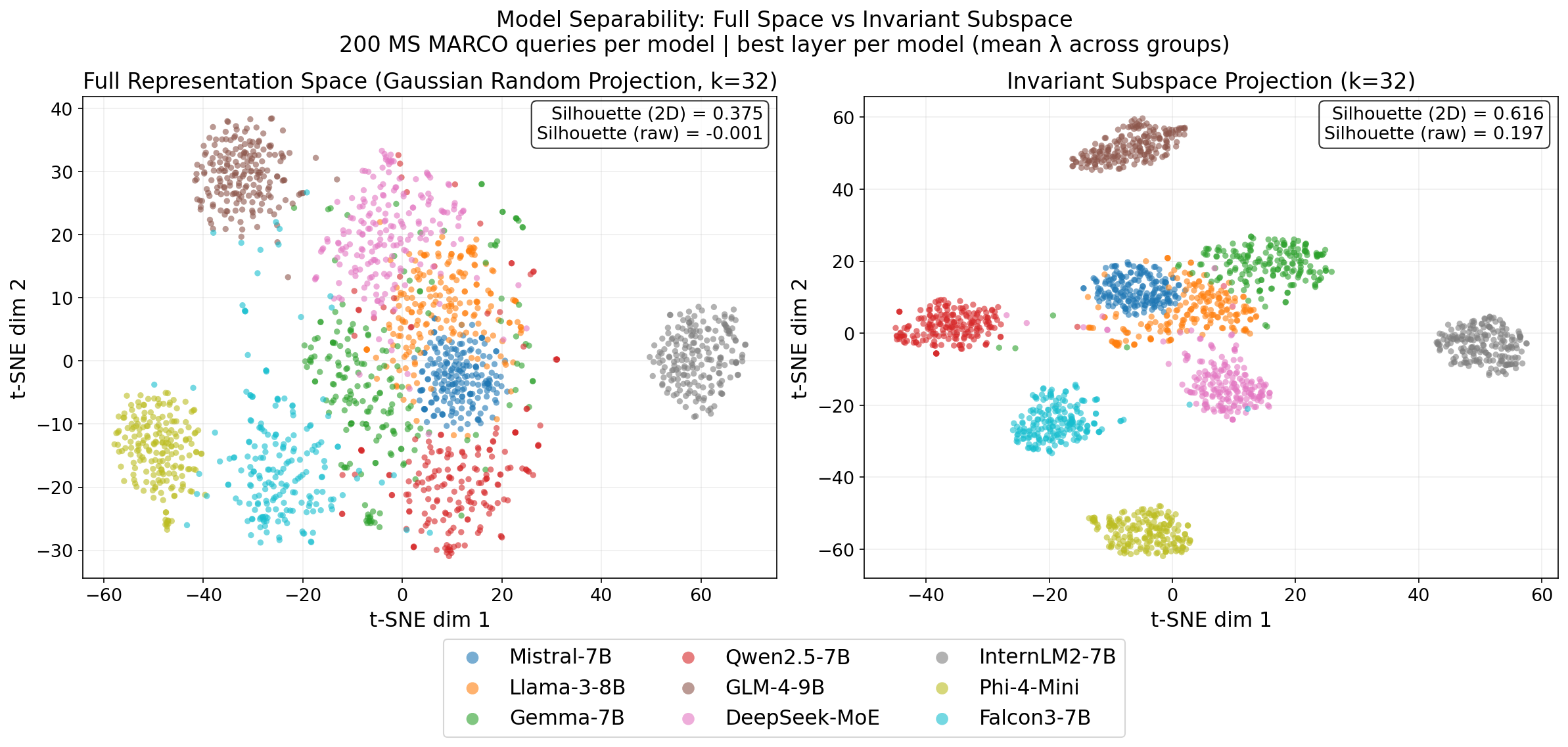}
    \vspace{-1.5em}
    \caption{Side-by-side t-SNE comparison of model separability using 200 MS MARCO queries per model ($k=32$). \textit{Left:} full representation space reduced via Gaussian random projection. \textit{Right:} invariant subspace projection. Silhouette scores are shown for both the 2D t-SNE output and the raw $k$-dimensional coordinates.}
    \label{fig:tsne_comparison}
    \vspace{-0.5em}
\end{figure}

Figure~\ref{fig:tsne_comparison} compares two conditions. The left panel reduces each model's raw $d$-dimensional hidden states to $k=32$ dimensions via Gaussian random projection, a distance-preserving linear transformation that treats all directions uniformly and introduces no semantic bias, before applying t-SNE. The right panel applies t-SNE directly to invariant subspace coordinates $c = U_{\mathrm{inv}}^\top h \in \mathbb{R}^{32}$, which already share a common dimensionality across models. To quantify separability beyond visual inspection, silhouette scores are computed both on the 2D t-SNE embeddings and on the original $k$-dimensional representations, with both values reported in the figure. The invariant projection exhibits consistently higher separability, indicating that the contrastive subspace concentrates model-discriminative structure while suppressing nuisance variation. This is consistent with the geometric formulation in Section~\ref{sec: find}, where invariant directions are selected to maximize semantic-changing variation relative to semantic-preserving variation.

\section{Broader Impacts}
\label{app:broader_impacts}

This work studies the internal representation geometry of large language models and proposes a framework for identifying invariant latent features that encode semantic information. As a primarily foundational contribution, the goal is to improve scientific understanding of how language models organize meaning and maintain consistency under paraphrastic variation.

A potential positive impact of this work lies in model transparency and interpretability. By providing a principled method to analyze invariant structure, the framework may support better diagnostic tools for understanding model behavior, detecting failure modes, and evaluating robustness under distribution shift. In addition, the use of invariant representations for model attribution may contribute to provenance tracking, intellectual property protection, and forensic analysis of generated content, which are increasingly important in the deployment of generative models.

At the same time, the ability to identify model-specific signatures from internal representations may introduce potential risks. Attribution techniques could be used to infer proprietary information about models or to reverse-engineer aspects of their internal structure without authorization. Furthermore, improved understanding of invariant features may enable adversaries to design attacks that specifically target or manipulate these representations, potentially affecting model reliability or enabling evasion of detection systems.

More broadly, as with many advances in representation learning, the proposed methods are dual-use. While they can enhance robustness, transparency, and accountability, they may also be misused in adversarial settings. We do not release model-specific fingerprints or tools intended for unauthorized identification, and we emphasize that responsible use of such techniques should follow appropriate legal and ethical guidelines.

Overall, we believe the primary impact of this work is to advance the scientific understanding of representation geometry in language models. We encourage future work to further study safeguards, robustness, and responsible deployment of techniques that analyze or utilize invariant latent structure.